\documentclass{article}

\usepackage[utf8]{inputenc}
\usepackage[english]{babel}
\usepackage{amsthm,amsmath,amsfonts,amssymb}
\usepackage[T1]{fontenc}
\usepackage{tikz}
\usetikzlibrary{shapes.geometric}
\usetikzlibrary{arrows.meta}
\usepackage{graphicx}
\usepackage{xcolor}
\usepackage{textcomp}
\usepackage{microtype}
\usepackage{graphicx}
\usepackage{subcaption}
\usepackage{booktabs} 

\usepackage{hyperref}

\usepackage{cleveref}

\usepackage[accepted]{icml2020}

\icmltitlerunning{Bootstrap Latent-Predictive Representations}

\newcommand{\Sspace}{\mathbb{S}}
\newcommand{\Ospace}{\mathbb{O}}
\newcommand{\Aspace}{\mathbb{A}}

\newcommand{\FSpace}{\mathbb{F}}
\newcommand{\GSpace}{\mathbb{G}}

\newcommand{\acronym}{PBL}
\newcommand{\technique}{Predictions of Bootstrapped Latents}

\definecolor{zesty_1}{HTML}{F5793A}
\definecolor{zesty_2}{HTML}{A95AA1}
\definecolor{zesty_3}{HTML}{85C0F0}
\definecolor{zesty_4}{HTML}{0F2080}

\tikzset{%
  stopGradTip/.tip={Bar[sep]Bar[sep=2]Triangle[angle=45:4pt]},
  gradTip/.tip={Triangle[angle=45:4pt]},
}

\begin{document}

\twocolumn[
\icmltitle{Bootstrap Latent-Predictive Representations \\for Multitask Reinforcement Learning}



\icmlsetsymbol{equal}{*}

\begin{icmlauthorlist}
\icmlauthor{Daniel Guo}{equal,DeepMindUK}
\icmlauthor{Bernardo Avila Pires}{equal,DeepMindUK}
\icmlauthor{Bilal Piot}{DeepMindUK}
\icmlauthor{Jean Bastien Grill}{DeepMindParis}
\icmlauthor{Florent Altch\'e}{DeepMindParis}
\icmlauthor{R\'emi Munos}{DeepMindParis}
\icmlauthor{Mohammad Gheshlaghi Azar}{DeepMindUK}
\end{icmlauthorlist}

\icmlaffiliation{DeepMindUK}{DeepMind, London, UK}
\icmlaffiliation{DeepMindParis}{DeepMind Paris, France}

\icmlcorrespondingauthor{Mohammad Gheshlaghi Azar}{mazar@google.com}

\icmlkeywords{Reinforcement Learning, POMDPs, Representation Learning, Predictive Representations}

\vskip 0.3in
]



\printAffiliationsAndNotice{\icmlEqualContribution} 

\begin{abstract}
Learning a good representation is an essential component for deep reinforcement learning (RL). 
Representation learning is especially important in multitask and partially observable settings where building a representation of the unknown environment is crucial to solve the tasks. 
Here we introduce \technique{} (\acronym{}), a simple and flexible self-supervised representation learning algorithm for multitask deep RL.
\acronym{} builds on multistep predictive representations of future observations, and focuses on capturing structured information about environment dynamics.
Specifically, \acronym{} trains its representation by predicting latent embeddings of future observations. These latent embeddings are themselves trained to be predictive of the aforementioned representations. These predictions form a bootstrapping effect, allowing the agent to learn more about the key aspects of the environment dynamics.
In addition, by defining prediction tasks completely in latent space, \acronym{} provides the flexibility of using multimodal observations involving pixel images, language instructions, rewards and more. 
We show in our experiments that \acronym{} delivers across-the-board improved performance over state of the art deep RL agents in the DMLab-30 and Atari-57 multitask setting. 
\end{abstract}

\section{Introduction}
Deep reinforcement learning (RL) has seen many successes in the recent years~\citep{mnih2015human,levine2016end,silver2017mastering,vinyals2019grandmaster}. However, there are still many improvements to be made in complex, multitask and partially observable environments. Previous work~\citep{mirowski2016learning, jaderberg2017reinforcement, hessel2019multi, gregor2019shaping} has demonstrated that augmenting deep RL agents with auxiliary tasks for representation learning helps improve RL task performance. 
Representation learning has also had broader impact in deep RL, as representations have been used to generate intrinsic motivation signal for exploration~\citep{houthooft2016vime, pathak2017curiosity, burda2019exploration, azar2019world, badia2020never} and as the basis for search models in planning agents~\citep{oh2017value, amos2018learning, hafner2019learning, schrittwieser2019mastering}.

A common  approach for representation learning is through training a (recurrent) neural network (RNN) to make predictions on future observations in a self-supervised fashion~\citep{oh2017value, amos2018learning, moreno2018neural, guo2018neural, hafner2019learning, schrittwieser2019mastering, gregor2019shaping}. Predictive models for representation learning  have been widely used as auxiliary tasks for deep RL~\citep{jaderberg2017reinforcement,oord2018representation,zhang2018solar,aytar2018playing,gregor2019shaping,song2019vmpo}. However, to learn a rich and useful representation, these methods may demand a high-degree of accuracy in multistep prediction of future observations. This degree of accuracy can be difficult to achieve in many problems, especially in partially observable and multitask settings, where uncertainty in the agent state and complex, diverse observations make prediction more challenging. In these settings, instead of trying to predict everything accurately, we would ideally want to predict a latent embedding of future observations that focuses on the key, structural information of the tasks.

In this paper, we introduce \technique{} (\acronym{}, ``pebble''), a new representation learning technique for deep RL agents. \acronym{} learns a representation of the history \citep[\emph{agent state}, ][]{sutton2018reinforcement} by predicting the future \emph{latent} embeddings of observations. The novel contribution is in how we train these latent embeddings, which we do by having the latent observation embedding at each timestep be predictive of the agent state at the same timestep. This leads to a bootstrapping effect, where the agent states learn to predict future latent observations, and future latent observations learn to predict the corresponding future agent states. 
As an example, consider a task where an agent must find a key and use it to unlock a door. 
All histories that lead to an unlocked door must also contain the observation of finding the key.
\acronym{} trains the latent embedding of observations of the unlocked door to predict the agent states that capture these histories, which means the latent embedding is encouraged to encode information pertaining to the key.
This example illustrate how representations trained with \acronym{} may capture contextual, structured information about the dynamics of the environment.
This is especially beneficial in settings with diverse and complex observations. 
Furthermore, \acronym{} allows for bootstrapping information from the far future, since the predicted agent state is also predictive of future. 
Finally, by making predictions solely in latent space, \acronym{} makes it easier to incorporate observations with multiple modalities such as images, voice and language instruction.

We evaluate \acronym{} in the challenging partially observable, multitask setting, where the task indices are kept hidden from the agent~\citep{brunskill2013sample}. As the benchmark, we use DMLab-30~\citep{beattie2016deepmind} as a standard partially observable and multitask RL domain, and compare the performance of various representation learning techniques when they are used as auxiliary representation learning tasks for the PopArt-IMPALA agent~\citep{hessel2019multi}, i.e., both RL and representation learning techniques train the same RNN. In particular we report the performance of \acronym{}, pixel control (state-of-the-art), DRAW~\citep{gregor2019shaping} and constrastive predictive coding (CPC)~\citep{oord2018representation} in this setting.
We show, in~\cref{sec:experiments}, that \acronym{}, is able to outperform all these methods both in terms of  final performance and sample efficiency. 
This result suggests that \acronym{} can be used as an alternative for the existing representation learning methods in the context of multitask deep RL.

\section{Background}
\paragraph{Partially Observable Environments.}
We model environments as partially observable Markov decision processes~\citep[POMDP, ][]{cassandra1994acting} where an agent does not directly perceive the underlying states but only a partial view of the underlying states. More formally, a POMDP is a tuple $(\Sspace, \Aspace, \Ospace, p_{\Sspace}, p_{\Ospace}, r, \gamma)$, where $\Sspace$ is the state space, $\Aspace$ the action space and $\Ospace$ the observation space. The transition probability kernel $p_{\Sspace}$ determines the probability of the next state $S_{t+1}$ given the current state and action $(S_t, A_t)$: $p_{\Sspace}(S_{t+1} = s'| S_t = s, A_t = a)$. The observation kernel $p_{\Ospace}$ dictates the observation process: $p_{\Ospace}(O_t = o|S_t = s)$.  $r\in\mathbb{R}^{\Sspace\times \Aspace}$ represents the  reward function and $\gamma$ is the discount factor.~\footnote{We note that the  multitask learning setting \citep{brunskill2013sample}, where the identity of task is kept hidden from the agent, is absorbed into the POMDP formulation, since the unobserved state can also index different per-task dynamics and reward functions.}

Policies dictate the action process: A policy is a mapping from sequences of observations and actions to a distribution over actions, that is, $\pi: \{ \Ospace \times (\Aspace \times \Ospace)^n : n \in \mathbb{N}\} \rightarrow \Delta(\Aspace)$, where $\Delta(\Aspace)$ denotes the space of action distributions.
For a fixed policy $\pi$ and for all $t\geq0$, we define recursively, the states $S^{\pi}_t$, the observations $O^{\pi}_t$, the histories $H^\pi_t$ and actions $A^\pi_t$ starting form  some initial distribution $\rho$ and following $\pi$:
\begin{align*}
\textit{Initialization:}&\left\{
    \begin{array}{ll}
          S^\pi_0 \sim \rho, \quad O^\pi_{0} \sim p_{\Ospace}(S^\pi_0), &\\
           H^\pi_0 \doteq O^\pi_0, \quad A^\pi_0 \sim \pi(H^\pi_0). &
    \end{array}
\right.
\\
\textit{Recurrence:}&\left\{
    \begin{array}{ll}
          S^\pi_{t+1} \sim p_{\Sspace}(S^\pi_t, A^\pi_t),\quad O^\pi_{t+1} \sim p_{\Ospace}(S^\pi_{t+1}),&\\
          H^\pi_{t+1} \doteq (H^\pi_{t}, A^\pi_{t}, O^\pi_{t+1}), &\\
          A^\pi_{t+1}\sim\pi(H^\pi_{t+1}). &
    \end{array}
\right.
\end{align*}
The \emph{partial history} $H^\pi_{t,k} \doteq (H^\pi_t, A^\pi_{t}, \ldots, A^\pi_{t+k-1})$ is formed by the history $H^\pi_t$ and the $k$ subsequent actions $A^\pi_t, \ldots, A^\pi_{t+k-1}$.

The RL~\citep{sutton2018reinforcement,szepesvari2010algorithms} problem in the POMDP setting is finding the policy that maximizes the expected and discounted sum of rewards:
\[
    \max_{\pi} E\left(\sum_{t=0}^\infty \gamma^t r(S^\pi_t, A^\pi_t)\right).
\]

A common strategy for RL in partially observable domains
is to compress the full history $H^{\pi}_t$ as the \emph{agent state} $B^{\pi}_t$ \citep{sutton2018reinforcement} using a neural network and solve
\[
    \max_{\pi} E\left(\sum_{t=0}^\infty \gamma^t r(B^\pi_t)\right),
\]
where $r( B^{\pi}_t) \doteq E(r(S^\pi_t, A^\pi_{t})| B^{\pi}_t) $. In this case the policy is often considered as a mapping from the agent state to distributions over actions.
To avoid clutter, we will drop the $\pi$ superscript from the POMDP variables.

\paragraph{Predictive Representations.}
Deep RL methods use neural networks to produce the agent state, and learn it from RL and self-supervised auxiliary tasks \citep{mnih2016asynchronous,jaderberg2017reinforcement}.
A common approach to learn agent states is to learn predictive representations \citep{oh2015action,jaderberg2017reinforcement,oh2017value,amos2018learning,guo2018neural,ha2018recurrent,moreno2018neural,oord2018representation,gregor2019shaping,hafner2019learning,schrittwieser2019mastering}.
One way to do so is to train representations to predict statistics of future observations conditioned on partial histories \citep{guo2018neural,gregor2019shaping}.
This technique uses RNNs to compress full histories $H_t$ as $B_t$ and partial histories $H_{t, k}$ as $B_{t,k}$, as inputs to auxiliary prediction tasks.

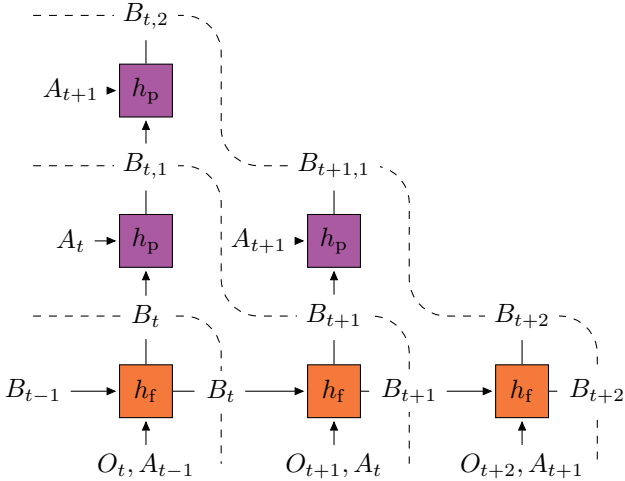
\begin{figure}[htb]
    \centering
    \begin{tikzpicture}[shorten >=1pt, node distance=\layersep]
        \tikzstyle{every pin edge}=[<-,shorten <=1pt]
        \tikzstyle{rnn}=[shape=rectangle,draw=black,fill=black!25,minimum size=2em]
        \draw [dashed,rounded corners=0.5cm] (-1.5, 1) -- (1.0, 1) -- (1.0, -1);
        \draw [dashed,rounded corners=0.5cm] (-1.5, 3) -- (1.0, 3) -- (1.0, 1) -- (3.5, 1) -- (3.5, -1);
        \draw [dashed,rounded corners=0.5cm] (-1.5, 5) -- (1.0, 5) -- (1.0, 3) -- (3.5, 3) -- (3.5, 1) -- (6.0, 1) -- (6.0, -1);
        \node[fill=white] (hist_0) at (-1.5, 0) {$B_{t-1}$};
        \node[rnn,fill=zesty_1] (cl_0) at (0.0, 0) {$h_{\mathrm{f}}$};
        \node[fill=white] (inp_0) at (0.0, -1) {$O_{t}, A_{t-1}$};
        \draw [-gradTip] (inp_0) -- (cl_0);
        \node[fill=white] (outp_0_0) at (0.0, 1) {$B_{t}$};
        \node[fill=white] (outp_0) at (1.0, 0) {$B_{t}$};
        \draw (cl_0) -- (outp_0_0);
        \draw (cl_0) -- (outp_0);
        \draw [-gradTip] (hist_0) -- (cl_0);
        \node[rnn,fill=zesty_2] (ol_0_1) at (0.0, 2) {$h_{\mathrm{p}}$};
        \node[fill=white] (inp_0_1) at (-1.0, 2) {$A_{t}$};
        \draw [-gradTip] (inp_0_1) -- (ol_0_1);
        \node[fill=white] (outp_0_1) at (0.0, 3) {$B_{t,1}$};
        \draw (ol_0_1) -- (outp_0_1);
        \draw [-gradTip] (outp_0_0) -- (ol_0_1);
        \node[rnn,fill=zesty_2] (ol_0_2) at (0.0, 4) {$h_{\mathrm{p}}$};
        \node[fill=white] (inp_0_2) at (-1.0, 4) {$A_{t+1}$};
        \draw [-gradTip] (inp_0_2) -- (ol_0_2);
        \node[fill=white] (outp_0_2) at (0.0, 5) {$B_{t,2}$};
        \draw (ol_0_2) -- (outp_0_2);
        \draw [-gradTip] (outp_0_1) -- (ol_0_2);
        \node[rnn,fill=zesty_1] (cl_1) at (2.5, 0) {$h_{\mathrm{f}}$};
        \node[fill=white] (inp_1) at (2.5, -1) {$O_{t+1}, A_{t}$};
        \draw [-gradTip] (inp_1) -- (cl_1);
        \node[fill=white] (outp_1_0) at (2.5, 1) {$B_{t+1}$};
        \node[fill=white] (outp_1) at (3.5, 0) {$B_{t+1}$};
        \draw (cl_1) -- (outp_1_0);
        \draw (cl_1) -- (outp_1);
        \draw [-gradTip] (outp_0) -- (cl_1);
        \node[rnn,fill=zesty_2] (ol_1_1) at (2.5, 2) {$h_{\mathrm{p}}$};
        \node[fill=white] (inp_1_1) at (1.5, 2) {$A_{t+1}$};
        \draw [-gradTip] (inp_1_1) -- (ol_1_1);
        \node[fill=white] (outp_1_1) at (2.5, 3) {$B_{t+1,1}$};
        \draw (ol_1_1) -- (outp_1_1);
        \draw [-gradTip] (outp_1_0) -- (ol_1_1);
        \node[rnn,fill=zesty_1] (cl_2) at (5.0, 0) {$h_{\mathrm{f}}$};
        \node[fill=white] (inp_2) at (5.0, -1) {$O_{t+2}, A_{t+1}$};
        \draw [-gradTip] (inp_2) -- (cl_2);
        \node[fill=white] (outp_2_0) at (5.0, 1) {$B_{t+2}$};
        \node[fill=white] (outp_2) at (6.0, 0) {$B_{t+2}$};
        \draw (cl_2) -- (outp_2_0);
        \draw (cl_2) -- (outp_2);
        \draw [-gradTip] (outp_1) -- (cl_2);

    \end{tikzpicture}
    \caption{Recurrent architecture for compressing partial histories.
    Networks used for processing observations and actions have been omitted, and dashed lines connect histories and partial histories aligned in time.} \label{fig:rnns}
\end{figure}

\Cref{fig:rnns} outlines the RNN architecture for compressing histories.
The horizontal direction shows the RNN $h_{\mathrm{f}}$ compressing full histories, with actions and observations as inputs.
The agent state is updated as $B_{t+1} \doteq h_{\mathrm{f}}(B_t, O_{t+1}, A_{t})$ (with $B_0$ equal zero).

The vertical direction shows the RNN $h_{\mathrm{p}}$ compressing partial histories, starting from agent states, and continuing with actions as the only inputs.
The state of $h_{\mathrm{p}}$ is updated as
\begin{align*}
    B_{t,1} &\doteq h_{\mathrm{p}}(B_t, A_t) \\
    B_{t,k+1} &\doteq h_{\mathrm{p}}(B_{t,k}, A_{t+k}),
\end{align*}

One can use the agent state $B_t$ as the basis (input) for the decision making of the RL algorithm \citep{jaderberg2017reinforcement,espeholt2018scalable,hessel2019multi}.
The partial history RNN $h_{\mathrm{p}}$ is only used for predictions about the future.

\section{\technique{} (\acronym{})}
\label{sec:technique}

Our method, \technique{} (\acronym{}, ``pebble''), consists of two auxiliary prediction tasks: 1) a forward, action-conditional prediction from compressed partial histories to future latent observations; and 2) a reverse prediction from latent observations to agent states.

The forward, action-conditional prediction task of \acronym{} consists of predicting latent embeddings, $Z_{t+k} = f(O_{t+k})$, of future observations, from a compressed partial history $B_{t, k}$:
\begin{align}
   E(Z_{t+k}|B_{t,k})&\approx E(Z_{t+k}|H_{t,k})
   \\
   &= E(Z_{t+k}|H_t, A_t, \ldots, A_{t+k-1}).\label{eq:futurePrediction}
\end{align}
To learn rich representations we make many predictions into the future---we take $k$ to range from one to a maximum index in the future, which we call the \emph{horizon}.

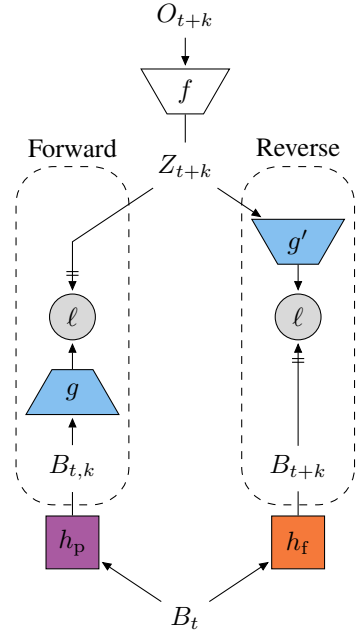
\begin{figure}[htb]
    \centering
    \begin{tikzpicture}[shorten >=1pt, node distance=\layersep]
        \tikzstyle{mlp}=[shape=trapezium,draw=black,fill=black!25,minimum size=17pt]
        \tikzstyle{conv}=[shape=rectangle,draw=black,fill=black!25,minimum size=17pt,rounded corners=.8ex]
        \tikzstyle{rnn}=[shape=rectangle,draw=black,fill=black!25,minimum size=2em]
        \tikzstyle{loss}=[shape=circle,draw=black,fill=black!15,minimum size=17pt,rounded corners=.8ex]
    
        \node[] (base_history) at (0, 0) {$B_t$};
        
        \node[rnn,fill=zesty_2] (h_p) at (-1.5, 1) {$h_{\mathrm{p}}$};
        \node[] (partial_history) at (-1.5, 2) {$B_{t,k}$};
        \node[mlp,fill=zesty_3] (g) at (-1.5, 3) {$g$};
        \node[loss] (forward_l) at (-1.5, 4) {$\ell$};
        \node[] (latent) at (0, 6) {$Z_{t+k}$};
        
        \node[mlp,fill=white,shape border rotate=180] (f) at (0, 7) {$f$};
        \node[] (o) at (0, 8) {$O_{t+k}$};
        
        \node[rnn,fill=zesty_1] (h_f) at (1.5, 1) {$h_{\mathrm{f}}$};
        \node[] (full_history) at (1.5, 2) {$B_{t+k}$};
        \node[loss] (reverse_l) at (1.5, 4) {$\ell$};
        \node[mlp,fill=zesty_3,shape border rotate=180] (g_prime) at (1.5, 5) {$g'$};

        \draw [-gradTip] (base_history) -- (h_p);
        \draw (h_p) -- (partial_history);
        \draw [-gradTip] (partial_history) -- (g);
        \draw [-gradTip] (g) -- (forward_l);
        \draw [-stopGradTip] (latent) -- (-1.5, 5) -- (forward_l);
        
        \draw [-gradTip] (o) -- (f);
        \draw (f) -- (latent);
        
        \draw [-gradTip] (base_history) -- (h_f);
        \draw (h_f) -- (full_history);
        \draw [-stopGradTip] (full_history) -- (reverse_l);
        \draw [-gradTip] (g_prime) -- (reverse_l);
        \draw [-gradTip] (latent) -- (g_prime);
        
        \node[] at (-1.5, 6.25) {Forward};
        \node[] at (1.5, 6.25) {Reverse};
        
        \draw [dashed,rounded corners=0.5cm] (-2.25, 1.5) -- (-2.25, 6) -- (-0.75, 6) -- (-0.75, 1.5) -- cycle;
        \draw [dashed,rounded corners=0.5cm] (2.25, 1.5) -- (2.25, 6) -- (0.75, 6) -- (0.75, 1.5) -- cycle;
    \end{tikzpicture}
    \caption{Architecture diagram for \acronym{}, forward prediction (left) and reverse prediction (right). Crossed arrows block gradients and $\ell(x, y) \doteq \| x - y \|^2_2$.}
    \label{fig:technique}
\end{figure}

To predict $Z_{t+k}$, we solve:
\begin{equation}
    \min_{h \in \mathbb{H}, g \in \GSpace} \sum_{t,k}\|g(B_{t,k}) - Z_{t+k}\|^2_2,
    \label{eq:forwardObjective}
\end{equation}
where $\mathbb{H}$ and $\GSpace$ are hypothesis spaces induced by neural networks, $g$ is a feed-forward neural network, and $h=(h_{\mathrm{f}}, h_{\mathrm{p}})$ are the RNN networks that compute $B_t$ and $B_{t,k}$.
We call this problem a \emph{forward prediction} (from compressed partial histories to latents), and its schematic is given in \cref{fig:technique}.

The choice of $Z_t$ is an essential part of our representation learning.
By picking $Z_t$ to be a \emph{latent} embedding of observations, we are able to learn meaningful observation features, as well as combine multiple modalities of observations (images and natural language sentences, for example) into a single representation.
Specifically, \acronym{} trains the latent embedding neural network $f$ by predicting the learned $B_{t}$ from $Z_{t}$. Concretely, we solve
\begin{equation}
    \min_{f \in \FSpace, g' \in \GSpace'} \sum_{t}\| g'(\overbrace{f(O_t)}^{Z_t}) - B_{t} \|^2_2,
    \label{eq:backwardObjective}
\end{equation}
where $\FSpace$ and $\GSpace'$ are hypothesis spaces induced by neural networks, and $g'$ is another feed-forward neural network.
We call this problem \emph{reverse prediction}, from latents to compressed histories, and its schematic is also given in \cref{fig:technique}.

Forward prediction trains agent states to predict future latent observations. Reverse prediction trains latent observations to predict immediate agent states. Together, they can form a cycle, bootstrapping useful information between compressed histories and latent observations.

Because of the bootstrapping effect, the reverse prediction has the potential to train meaningful latent embeddings of observations that encode structural information about the paths that the agent takes to reach these observations.  Furthermore, since we are training agent states to be predictive, this bootstrapping effect may also bootstrap useful information from far into the future. 

Note that both forward and reverse predictions are one-way, i.e., we do not pass gradients into the targets of the predictions. In practice, we optimize both objectives together, as they optimize disjoint sets of parameters. A natural question that arises is what makes the bootstrapped predictions work, as it seems like there is the possibility that the prediction tasks collapse into trivial solutions. We hypothesize that due to forward and reverse predictions being one-way, the training dynamics actively move away from the trivial solution. Initially, the latent embedding of observations and agent states are the output of randomly initialized neural networks, which can be considered as random, non-linear projections. The forward prediction encourages the agent state to move away from collapsing in order to accurately predict future random projections of observations. Similarly, the reverse prediction encourages the latent observation away from collapsing in order to accurately predict the random projection of a full history. As we continue to train forward and reverse predictions, this seems to result in a virtuous cycle that continuously enriches both representations. In \cref{sec:collapse}, we empirically analyze this issue.

\section{Experiments}
\label{sec:experiments}
In this section, we present our main experimental results (\cref{sec:dmlab}) where we show that \acronym{}, when used as the auxiliary task to shape the representation, outperforms other representation learning auxiliary tasks in DMLab 30 benchmark.
Then we empirically investigate the scalability of \acronym{} with the horizon of prediction and the effect of the reverse prediction in \cref{sec:prp}, the stability of \acronym{} in \cref{sec:collapse}, the neural network architecture choice in \cref{sec:architecture}, and what is captured by the learned representation in \cref{sec:decode}.
We discuss implementation details in \cref{sec:implementationOverview}. 
Additionally, we performed experiments in the standard Atari-57 domain \citep{bellemare2013arcade}, which we report in \cref{sec:atari}.

\subsection{DMLab 30}
\label{sec:dmlab}
We evaluated \acronym{} in the DMLab 30 task set \citep{beattie2016deepmind}. 
In this benchmark the agent must perform navigation and puzzle-solving tasks in 3D environments. There are a total of 30 tasks, each with a number of procedural variations over episodes.
In the  multitask setup of DMLab 30 each episode may come from a different task, and the agent \emph{does not} receive the task identifier as an input. The only input the agent receives is an image of the first-person view of the environment, a natural language instruction (in some tasks), and the reward.

We compared \acronym{} with different representation learning techniques, when they are used as auxiliary tasks for the deep RL agent. Pixel control~\citep{jaderberg2017reinforcement} is the current state-of-the-art self-supervised representation learning technique used in DMLab 30, having been applied to PopArt-IMPALA~\citep{hessel2019multi} and, more recently, V-MPO~\citep{song2019vmpo}.
CPC has been successful for representation learning in different settings~\citep{oord2018representation}. One advantage of CPC is its simplicity; it is lightweight and easy to incorporate in different applications.
Simcore DRAW \citep{gregor2019shaping} has achieved strong performance in some of the DMLab 30 tasks in the single-task regime. It is also a good method to compare with \acronym{}, since it learns the  representation with multistep predictions of the future.
In contrast to CPC (which uses a contrastive approach for representation learning), Simcore DRAW uses a strong generative model of frames \citep{gregor2015draw}.

All the representation learning techniques have been combined with PopArt-IMPALA \citep{hessel2019multi} as the base RL algorithm.
PopArt-IMPALA has been shown to perform well in DMLab 30, where the use of PopArt makes the agent suitable for handling different reward scales across tasks. For \acronym{}, Simcore DRAW, and CPC, we predict up to 20 steps into the future.

The most common performance measure for DMLab 30 is the mean over tasks of the capped human normalized score.
The human normalized score of an agent on a task $i$ is calculated as $100 \cdot (a_i - u_i)/(h_i - u_i)$, where $a_i$ is the agent's expected return on that task, $u_i$ is the expected return of an agent that selects actions uniformly at random, and $h_i$ is the expected return of a human performing the task.
Reference values for the random and human scores can be found in the work of \citet{espeholt2018scalable}.
The \emph{capped} human normalized score of an agent on a task is the human normalized score capped at $100$.
Therefore, the mean capped human normalized emphasizes performance gains toward attaining human performance across tasks and disregards improvements that are superhuman.
In this work, we primarily report this score when aggregating results across tasks, but the mean human normalized score can be found in \cref{sec:extra}.

\begin{figure}[htb]
    \centering
    \includegraphics[width=0.45\textwidth]{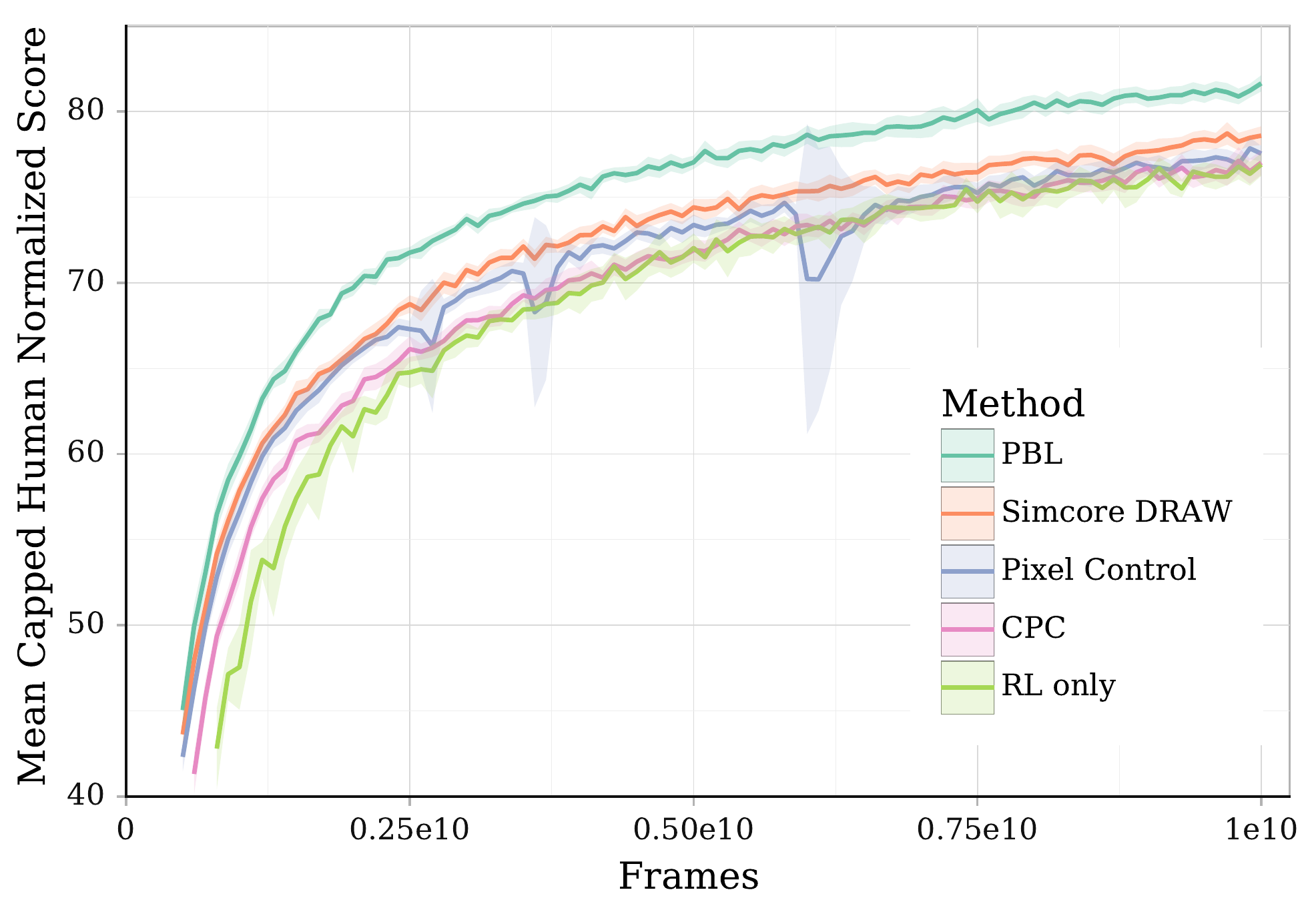}
    \caption{Mean capped human normalized score for compared methods.}
    \label{fig:comparison}
\end{figure}

\Cref{fig:comparison} shows the mean capped human normalized score for \acronym{} and our baselines, averaged across 16 independent runs, and with 95\% confidence intervals.
\Cref{app:plotting} gives additional details on the plotting protocol for our experiments.
We see that \acronym{} outperforms all other methods throughout the whole training process by a significant margin. Interestingly, Simcore DRAW also outperforms the standard pixel control, albeit to a lesser extent. This may be explained by the intuition that pixel-based representation learning may struggle with the large variety of tasks. The large diversity of images in multitask environments greatly increases the challenge of pixel prediction. Also, losses in pixel space can overlook details that are small but important for solving the tasks. Similarly, diverse images produced by different tasks are easy to distinguish from each other, meaning that CPC may not require a rich and informative representations to distinguish between the positive and negative examples, resulting in worse overall performance.

\begin{figure}[htb]
    \centering
    \includegraphics[width=0.45\textwidth]{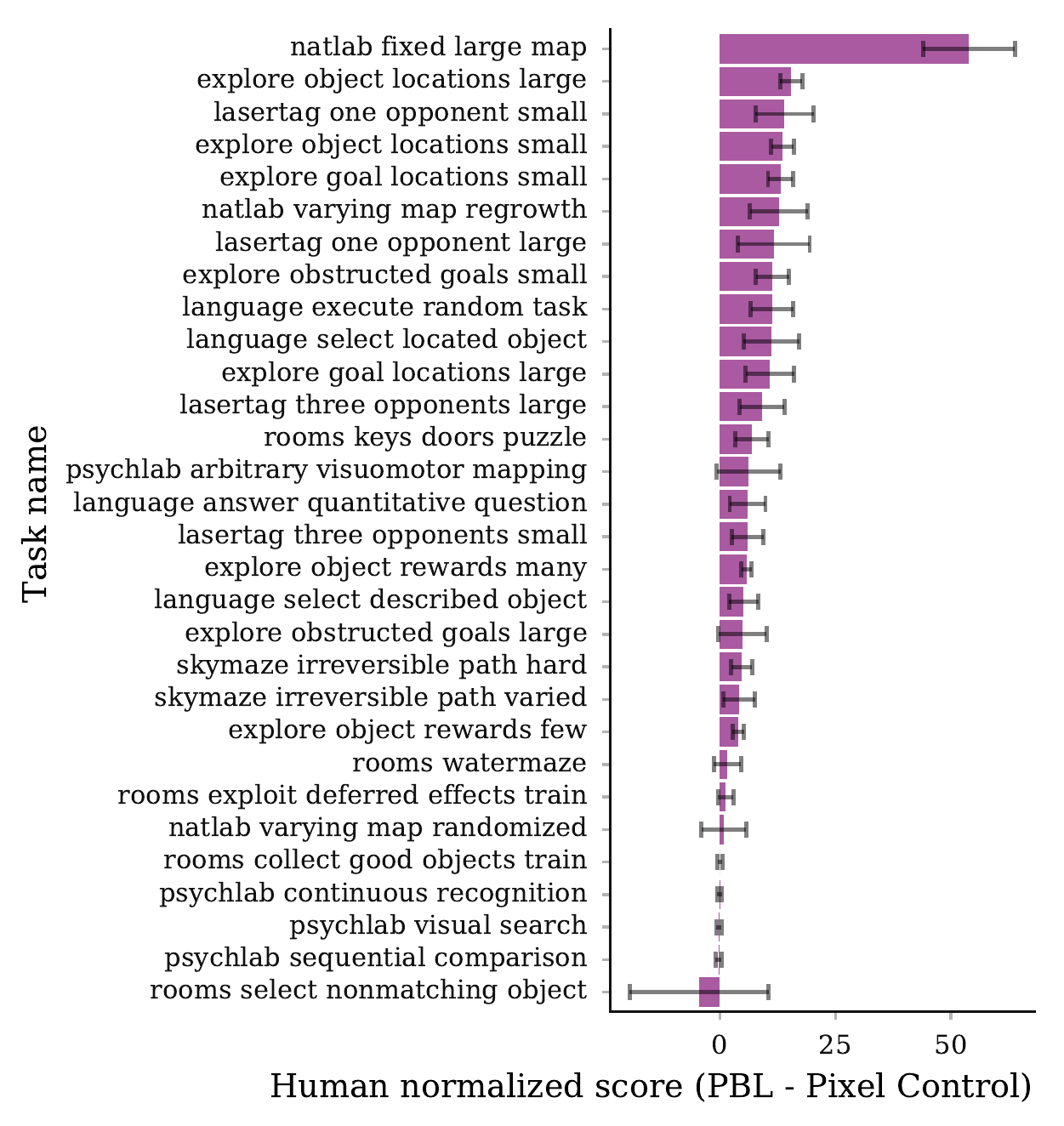}
    \caption{Difference in Human Normalized Score per Task between \acronym{} and pixel control.
    Error bars denote 95\% confidence intervals.}
    \label{fig:breakdown}
\end{figure}

We show a per-task breakdown of \acronym{} final performance relative to pixel control in \cref{fig:breakdown}.
We see that \acronym{} is able to boost performance across the majority of the tasks, while achieving parity with pixel control on the rest of the tasks. This result shows that the representation learned by \acronym{} encodes meaningful information which is useful to improve the performance of the agent across the board.

\subsection{Scalability with the Horizon of Prediction}
\label{sec:prp}
\begin{figure}[htb]
    \centering
    \includegraphics[width=0.45\textwidth]{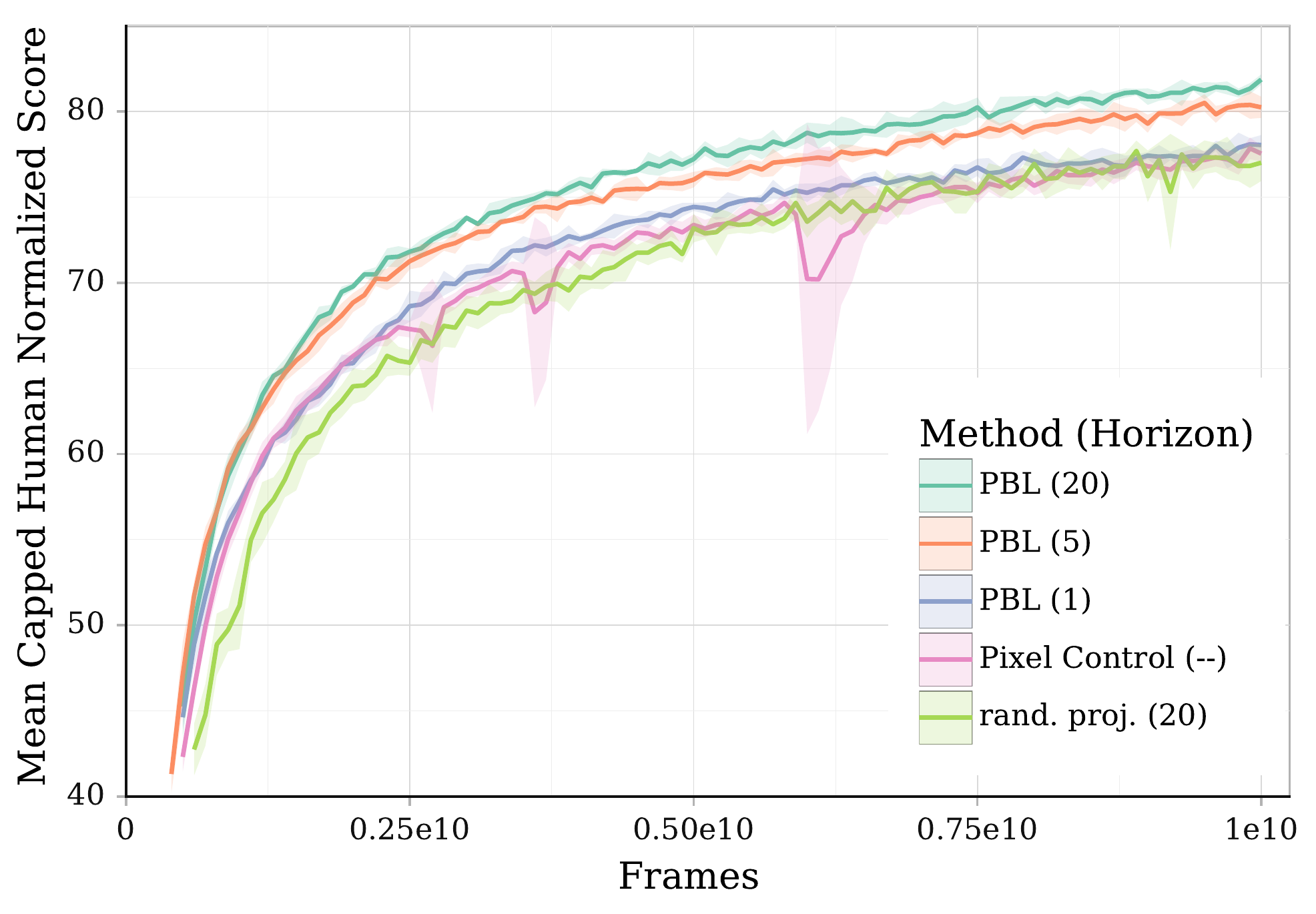}
    \caption{\acronym{} Performance Across Forward Prediction Horizon, compared to pixel control and random projection.}
    \label{fig:scalability}
\end{figure}

To have a  better understanding of the effect of horizon on the performance of \acronym{} we tried different horizon lengths for forward prediction in DMLab-30. \Cref{fig:scalability} shows that indeed, the performance improves monotonically as we predict farther into the future. However, as we increase the horizon, we get diminishing performance gains. Interestingly, predicting only one step into the future matches the performance of pixel control. This shows that multistep prediction enables \acronym{} to encode crucial information  for solving the tasks in its representation. 
To gain a better understanding on the effect of the prediction horizon,  we investigate the case where we disable reverse prediction (see \cref{fig:technique}), which means the latents are mere random projections of the observations.
Interestingly, we found that increasing the horizon makes no difference to performance, so we only report the performance for horizon $20$ in \cref{fig:scalability}. 
Overall, it is slightly worse than pixel control during training, but achieves the same final performance. These results emphasize that it is not enough to only try to predict farther into the future, but that it is crucial to predict meaningful latents about the future. This suggests that reverse prediction in \acronym{} is able to learn a meaningful representation of future observations that does extends far into the future.

\subsection{Stability of \acronym{}}
\label{sec:collapse}
From our experimental results, \acronym{} is perfectly stable, and indeed does not collapse into a trivial solution. To further investigate this issue, we have tested a variant of \acronym{}  combined with random projection---we predict two separate latent observation embeddings where one is learned as usual and the other is a random projection. This variant grounds the forward prediction, which explicitly prevents collapsing. Interestingly, this variant achieves identical performance to \acronym{}. This suggests that the training dynamics coupled with randomly initialized neural networks provide enough signal at the beginning to move away from collapse. As mentioned in \cref{sec:technique}, we suspect that the forward and reverse predictions seem to form a virtuous cycle that continuously enriches both the compressed full history and the latent observations, instead of driving them to a trivial solution.

\subsection{Architecture Choice}
\label{sec:architecture}

In our experiments, we used a variant of the architecture used by \citet{gregor2019shaping,song2019vmpo} (see \cref{tab:mainArchitecture} in \cref{sec:implementation} for full details).
We compared the base RL method (PopArt-IMPALA with no auxiliary tasks) with this larger architecture, against the same method with the architecture used by \citet{hessel2019multi}.

\begin{figure}[htb]
    \centering
    \includegraphics[width=0.45\textwidth]{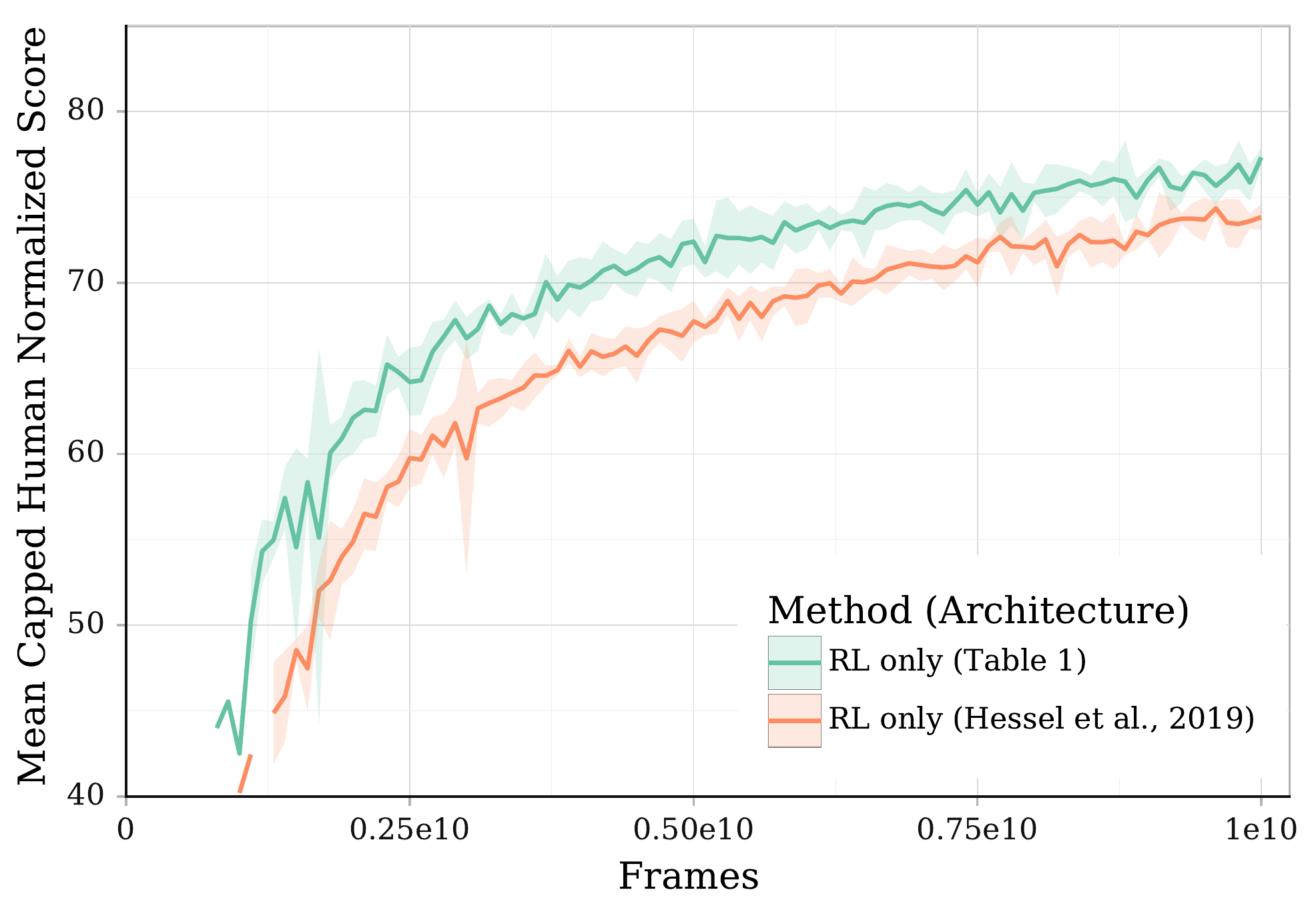}
    \caption{Mean capped human normalized score for compared architectures.}
    \label{fig:comparison-impala-capped}
\end{figure}

\Cref{fig:comparison-impala-capped} shows the mean capped human normalized score of the base RL method with the two architectures, across four different independent runs.
We chose to adopt the larger architecture because it gives a significant boost in overall performance.
\subsection{Atari-57}
\label{sec:atari}
We also ran~\acronym{} along with pixel control, CPC and RL-only on the Atari-57~\citep{bellemare2013arcade} in a \emph{multitask} setup \citep{espeholt2018scalable,hessel2019multi}. The main goal of this experiment is to show that \acronym{} is a general representation approach that can be useful in benchmarks other than DMLab 30. 
We choose Atari-57 since it is a standard benchmark in deep RL.

The result from  Simcore DRAW is omitted as it is designed to output RGB images in the DMLab 30 format, and is incompatible with standard stacked, greyscale Atari frames format. We only changed two hyperparameters from DMLab 30: one is the entropy cost, which is now set to $0.01$; the second is the unroll length, which is set to $30$. In~\citet{espeholt2018scalable}, the unroll length was set to $20$, but we increased it to $30$ which facilitates multistep prediction in the case of~\acronym{} and CPC. We also clip the rewards to be between $-1$ and $1$ which is standard for Atari-57 benchmark \citet{espeholt2018scalable}.

\begin{figure}[!h]
    \centering
    \includegraphics[width=8cm]{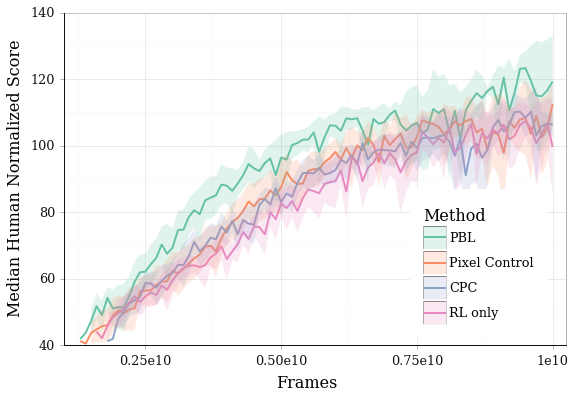}
    \caption{Median human normalized score for compared methods on Atari57.}
    \label{fig:atarimedian}
\end{figure}

\Cref{fig:atarimedian} shows the human normalized score (median across all 57 levels) for the compared approaches.
We see that \acronym{} is able to improve the overall performance in this benchmark.
Interestingly, it seems that adding auxiliary tasks for representation learning only offers at most minimal improvements in performance. 
This may be due to Atari being essentially a fully observable domain as opposed to the partially observable nature of DMLab.

\Cref{tab:finalAtariPerformance} shows the performance of different methods across the 57 Atari tasks with $95\%$ confidence intervals.
\acronym{} outperforms the other methods, including pixel control, for seven tasks.
In the other tasks, no method statistically outperforms \acronym{}.

\subsection{Decoding the \acronym{} Representation}
\label{sec:decode}
In this section, we try to shed some light on what information is being learned by the representation trained with \acronym{}. We designed a simple DMLab-like 3D environment in which the agent is in a small room. There is a single red cube that spawns in a random location in the room. We use a fixed, uniformly random policy and only train the representation without any reinforcement learning.

\begin{figure}[htb]
    \centering
    \includegraphics[width=0.3\textwidth]{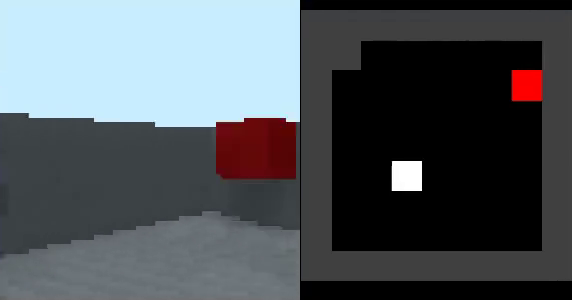}
    \caption{3D room example: Agent's first-person view (left) and top-down grid-view indicating the object position (right).}
    \label{fig:3dRoom}
\end{figure}

\begin{figure}[htb]
    \centering
    \includegraphics[width=0.44\textwidth]{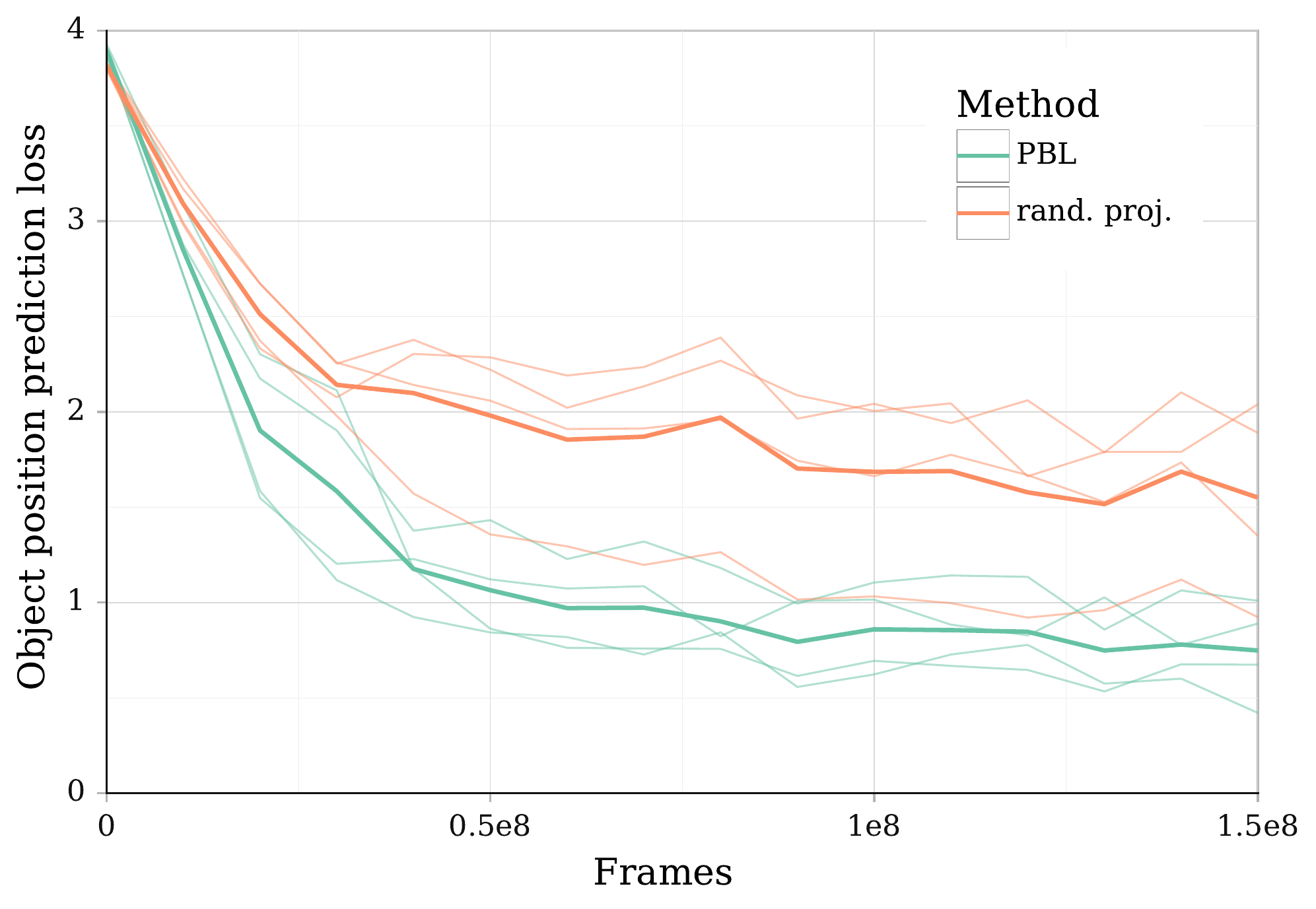}
    \caption{Object position prediction loss for random projection vs \acronym{}. Light lines denote different independent runs.}
    \label{fig:3dRoom-loss}
\end{figure}

\begin{figure}[htb]
    \centering
    \subcaptionbox{RP: Episode Middle}
    {\includegraphics[width=0.15\textwidth]{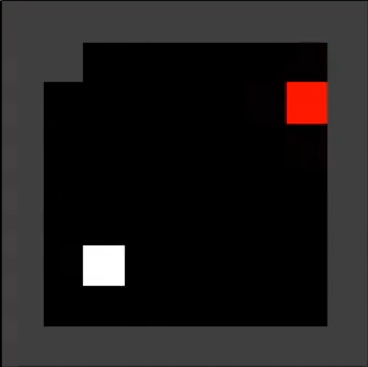}}
    \quad
    \subcaptionbox{RP: Episode End}
    {\includegraphics[width=0.15\textwidth]{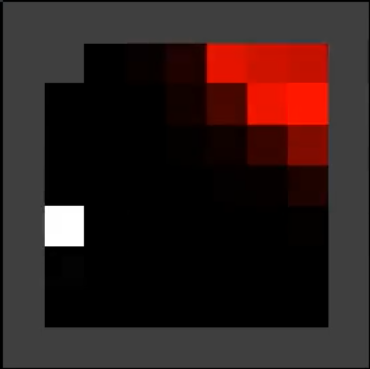}}
    \quad
    \subcaptionbox{\acronym{}: Episode Middle}
    {\includegraphics[width=0.15\textwidth]{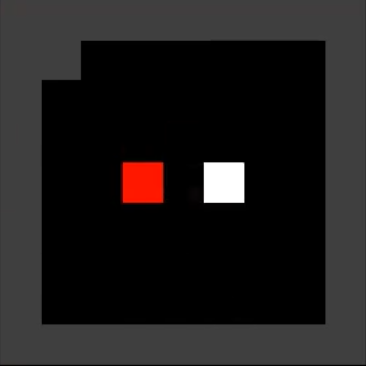}}
    \quad
    \subcaptionbox{\acronym{}: Episode End}
    {\includegraphics[width=0.15\textwidth]{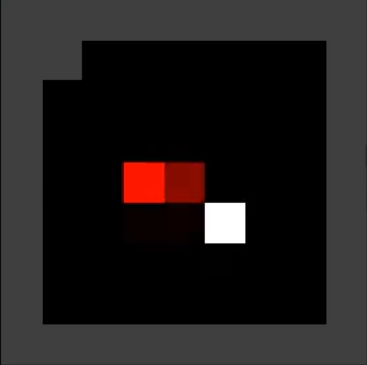}}
    \caption{Predicted top-down grid-view for random projection (top row) vs \acronym{} (bottom row)}
    \label{fig:3dRoom-comparison}
\end{figure}

Our criteria for a good representation is one that is able to encode the position of the red cube once the agent has seen it. To measure this, we train (in tandem with the agent's representation) a predictor from the agent's representation to the indicator of the position of the object in a top-down view of the environment.
This position predictor does not affect the agent's representation (that is, no gradients pass from the predictor to the representation)---this type of predictor is known as this is a probe \citep{amos2018learning} or a glass-box evaluator \citep{guo2018neural}. \Cref{fig:3dRoom} shows a first-person view of the room and a top-down grid-view of the environment indicating the ground-truth position of the object.
We compared \acronym{} against the random projection version where reverse prediction is disabled. 
\Cref{fig:3dRoom-loss} shows the loss for the object position predictor between these two methods, and clearly indicates that \acronym{} is learning a representation that is significantly better at encoding the position of the object. 
Taking a qualitative look at what happens during individual episodes, \Cref{fig:3dRoom-comparison} shows snapshots of the position prediction in the middle and at the end of the episode. It looks like all methods are able to encode the position of the red cube shortly after seeing it, but \acronym{} is much better at remembering the position much later on after the agent has looked away. This shows an intimate connection between predicting the future and effectively compressing the past history---if the agent can predict that at some point far in the future, it will turn around and see the object again, then that means the representation has remembered the position of the object.
\subsection{Implementation Overview}
\label{sec:implementationOverview}
In this section, we give an overview of the most relevant implementation details. A full account of our method, with details on architecture choices and parameters, can be found in \cref{sec:implementation}.

For the agent architecture (used in all methods), we use the same as \citet{hessel2019multi} except for the following changes: 1) We increased the LSTM from one 256 unit hidden layer to two 512 unit hidden layers; 2) We changed the linear policy and value heads to have each an additional 512 unit hidden layer. This is in fact akin to \citet{gregor2019shaping} but with wider layers. We duplicate the LSTM architecture to use for the partial history unroll. Our larger architecture resulted in strictly better performance for all methods.

For \acronym{}, we duplicate the agent's observation-processing architecture to use for the latent embedding ($f$ in \cref{fig:technique}). For the forward and reverse prediction networks ($g$ and $g'$ in \cref{fig:technique}) we use MLPs with 2 hidden layers of 512 units each. 

Most hyperparameters (e.g. PopArt hyperparameters) match the ones in \citet{hessel2019multi}. However, one significant difference is that we do not use population-based training (PBT). Instead, we picked the optimizer parameters and entropy penalty that worked best for the base RL agent (without any representation learning). For all representation learning methods, the only hyperparameter we tuned was the scale of the representation loss relative to the RL loss.

Computationally, predicting every future latent from one step ahead to the horizon from every timestep can be very expensive. We mitigate this issue by subsampling from which timesteps we do the forward prediction, and which future latents we predict \citep{gregor2019shaping}. This does not change the objective, and in practice can greatly speed up computation with minimal loss of performance.

\section{Related Work}
\label{sec:related}

\citet{littman2002predictive,singh2003learning} introduced predictive state representations (PSRs) and an algorithm for learning such representations.
PSRs are akin to action-conditional predictions of the future where the prediction tasks are indicators of events on the finite observation space.

The power of PSRs is the ability to act as a compact replacement for the state that can be constructed from observable quantities.
That is, PSRs are compact sufficient statistics of the history.
Inspired by this line of research, 
we focus on learning neural representations that are predictive of observations multiple timesteps into the future, conditioned on the agent's actions.

A number of other works investigated learning predictive representations (with action-conditional predictions of the future) to improve RL performance.
\citet{oh2015action} use generated observations in addition to the actions when compressing partial histories.
\citet{amos2018learning} learned representations by maximizing the likelihood of future proprioceptive observations using a PreCo model, which interleaves an action-dependent (predictor) RNN with an observation dependent (corrector) RNN for compressing full histories, and uses the predictor RNN for the action-only sequences in partial histories (\emph{cf.}~\cref{fig:rnns}, where one RNN is used for the full histories, and another for the partial histories).
\citet{oh2017value,schrittwieser2019mastering} learn to predict the base elements needed for Monte Carlo tree search conditioned on actions \citep{schrittwieser2019mastering} or options \citep{oh2015action}: Rewards, values, logits of the search prior policy (in the case of \citealp{schrittwieser2019mastering}), and termination/discount (in the case of \citealp{oh2017value}).
\citet{guo2018neural,moreno2018neural,gregor2019shaping} used an architecture similar to \cref{fig:rnns} with a particular focus on learning representations of belief state.
\citet{ha2018recurrent,hafner2019learning} used variational autoencoders \citep[VAEs, ][]{kingma2013auto,gregor2015draw} to shape the full history RNN representation, and trained latent models from the partial histories to maximize likelihood of the respective full histories. The Simcore DRAW technique \citep{gregor2019shaping} is a demonstrably strong VAE-based representation learning technique for single task deep RL.
It uses the same architecture as \cref{fig:rnns} to compress histories, but trains $h_f$ and $h_p$ by trying to autoencode observations.
Taking $Z'_{t+k} \sim P_{\phi}(Z'_{t+k}|O_{t+k})$ to be the DRAW latent with parameters $\phi$, the technique  maximizes the (parametric) likelihood $P_{\phi}(O_{t+k}|Z'_{t+k})$ subject to the posterior $P_\phi(Z'_{t+k}|O_{t+k})$ being close (in KL divergence) to the prior $P_\phi(Z'_{t+k}|B^\theta_{t+k})$.

\citet{jaderberg2017reinforcement} introduced pixel control as a representation learning task for reinforcement learning.
This technique applies Q-learning for controlling the change in pixels, which can be considered as a technique for learning predictive representations based on future pixel changes.

\citep[CPC][]{oord2018representation} is a powerful and lightweight alternative to generative models in representation learning.
Similar to \acronym{}, CPC operates in latent space.
In contrast to our approach, which is based on predicting latent embeddings, CPC uses a contrastive approach to representation learning.
(For an overview of CPC, see \cref{app:CPC}).

\section{Conclusion}

We considered the problem of representation learning for deep RL agents in partially observable, multitask settings.
We introduced \technique{} (\acronym{}), a representation learning technique that provides a novel way of learning meaningful future latent observations through bootstrapped forward and reverse predictions. We demonstrated that \acronym{} outperforms the state-of-the-art representation learning technique in DMLab-30.

Our results show that agents
in multitask setting
significantly benefit from learning to predict a \emph{meaningful} representation of the future in response to its actions.
This is highlighted in the fact that when we simply used random projections of future observations,
there was no benefit in trying to predict farther into the future.
On the other hand, by learning meaningful latent embeddings of observations, \acronym{} continued to improve as we increased the horizon.
Going forward, a promising direction would be to investigate ways to predict more relevant and diverse information, over long horizons.

Latent-space prediction is an  easy way to combine multiple observation modalities.
It would be worth seeing how far we can push the quality of learned representations by incorporating richer sources of observation.
To name a few, proprioception in robotics and control tasks \citep{levine2016end,tassa2018deepmind}, more sophisticated natural language, and even other sensory inputs like touch, smell, sound and temperature.   
In fact, the ability to learn representations from multimodal data makes \acronym{} a good candidate for evaluation in other machine learning domains, beyond RL, such as the settings considered by \citet{oord2018representation}.

A promising extension to \acronym{} may be to generalize the forward and reverse predictions from learning mean predictions (with squared loss) to learning generative models of latents. While this may not necessarily help in DMLab-30 where dynamics are deterministic, it may help in other settings where the dynamics are stochastic.
Having a generative model may also be used to generate rollouts for Monte Carlo planning \citep[similar to ][]{schrittwieser2019mastering}.

We think it would be fruitful to evaluate \acronym{} in a transfer learning setting.
Since \acronym{} improves performance virtually across all DMLab-30 tasks, it may have learned a representation that better captures the shared, latent structure across tasks, with the potential to generalize to new unseen tasks

Interesting future work also includes a more in-depth study of what predictive representations learn to encode, perhaps using a glass-box approach \citep{guo2018neural, amos2018learning}.
In particular, it is worth finding out how much information about the POMDP \emph{state} is captured and preserved by these representations.

\bibliographystyle{icml2020}
\bibliography{references}

\begin{thebibliography}{43}
\providecommand{\natexlab}[1]{#1}
\providecommand{\url}[1]{\texttt{#1}}
\expandafter\ifx\csname urlstyle\endcsname\relax
  \providecommand{\doi}[1]{doi: #1}\else
  \providecommand{\doi}{doi: \begingroup \urlstyle{rm}\Url}\fi

\bibitem[Abadi et~al.(2015)Abadi, Agarwal, Barham, Brevdo, Chen, Citro,
  Corrado, Davis, Dean, Devin, Ghemawat, Goodfellow, Harp, Irving, Isard, Jia,
  Jozefowicz, Kaiser, Kudlur, Levenberg, Man\'{e}, Monga, Moore, Murray, Olah,
  Schuster, Shlens, Steiner, Sutskever, Talwar, Tucker, Vanhoucke, Vasudevan,
  Vi\'{e}gas, Vinyals, Warden, Wattenberg, Wicke, Yu, and
  Zheng]{abadi2015tensorflow}
Abadi, M., Agarwal, A., Barham, P., Brevdo, E., Chen, Z., Citro, C., Corrado,
  G.~S., Davis, A., Dean, J., Devin, M., Ghemawat, S., Goodfellow, I., Harp,
  A., Irving, G., Isard, M., Jia, Y., Jozefowicz, R., Kaiser, L., Kudlur, M.,
  Levenberg, J., Man\'{e}, D., Monga, R., Moore, S., Murray, D., Olah, C.,
  Schuster, M., Shlens, J., Steiner, B., Sutskever, I., Talwar, K., Tucker, P.,
  Vanhoucke, V., Vasudevan, V., Vi\'{e}gas, F., Vinyals, O., Warden, P.,
  Wattenberg, M., Wicke, M., Yu, Y., and Zheng, X.
\newblock {TensorFlow}: Large-scale machine learning on heterogeneous systems,
  2015.
\newblock Software available from tensorflow.org.

\bibitem[Amos et~al.(2018)Amos, Dinh, Cabi, Roth{\"{o}}rl, Colmenarejo, Muldal,
  Erez, Tassa, de~Freitas, and Denil]{amos2018learning}
Amos, B., Dinh, L., Cabi, S., Roth{\"{o}}rl, T., Colmenarejo, S.~G., Muldal,
  A., Erez, T., Tassa, Y., de~Freitas, N., and Denil, M.
\newblock Learning awareness models.
\newblock In \emph{6th International Conference on Learning Representations,
  {ICLR} 2018, Vancouver, BC, Canada, April 30 - May 3, 2018, Conference Track
  Proceedings}, 2018.

\bibitem[Aytar et~al.(2018)Aytar, Pfaff, Budden, Paine, Wang, and
  de~Freitas]{aytar2018playing}
Aytar, Y., Pfaff, T., Budden, D., Paine, T.~L., Wang, Z., and de~Freitas, N.
\newblock Playing hard exploration games by watching {YouTube}.
\newblock In \emph{Advances in Neural Information Processing Systems 31: Annual
  Conference on Neural Information Processing Systems 2018, NeurIPS 2018, 3-8
  December 2018, Montr{\'{e}}al, Canada}, pp.\  2935--2945, 2018.

\bibitem[Azar et~al.(2019)Azar, Piot, Pires, Grill, Altch{\'e}, and
  Munos]{azar2019world}
Azar, M.~G., Piot, B., Pires, B.~A., Grill, J.-B., Altch{\'e}, F., and Munos,
  R.
\newblock World discovery models.
\newblock \emph{arXiv preprint arXiv:1902.07685}, 2019.

\bibitem[Beattie et~al.(2016)Beattie, Leibo, Teplyashin, Ward, Wainwright,
  Küttler, Lefrancq, Green, Valdés, Sadik, Schrittwieser, Anderson, York,
  Cant, Cain, Bolton, Gaffney, King, Hassabis, Legg, and
  Petersen]{beattie2016deepmind}
Beattie, C., Leibo, J.~Z., Teplyashin, D., Ward, T., Wainwright, M., Küttler,
  H., Lefrancq, A., Green, S., Valdés, V., Sadik, A., Schrittwieser, J.,
  Anderson, K., York, S., Cant, M., Cain, A., Bolton, A., Gaffney, S., King,
  H., Hassabis, D., Legg, S., and Petersen, S.
\newblock {DeepMind} lab, 2016.

\bibitem[Bellemare et~al.(2013)Bellemare, Naddaf, Veness, and
  Bowling]{bellemare2013arcade}
Bellemare, M.~G., Naddaf, Y., Veness, J., and Bowling, M.
\newblock The arcade learning environment: An evaluation platform for general
  agents.
\newblock \emph{Journal of Artificial Intelligence Research}, 47:\penalty0
  253--279, 2013.

\bibitem[Brunskill \& Li(2013)Brunskill and Li]{brunskill2013sample}
Brunskill, E. and Li, L.
\newblock Sample complexity of multi-task reinforcement learning.
\newblock In \emph{Proceedings of the Twenty-Ninth Conference on Uncertainty in
  Artificial Intelligence}, UAI’13, pp.\  122–131, Arlington, Virginia,
  USA, 2013. AUAI Press.

\bibitem[Burda et~al.(2019)Burda, Edwards, Storkey, and
  Klimov]{burda2019exploration}
Burda, Y., Edwards, H., Storkey, A.~J., and Klimov, O.
\newblock Exploration by random network distillation.
\newblock In \emph{7th International Conference on Learning Representations,
  {ICLR} 2019, New Orleans, LA, USA, May 6-9, 2019}, 2019.

\bibitem[Cassandra et~al.(1994)Cassandra, Kaelbling, and
  Littman]{cassandra1994acting}
Cassandra, A.~R., Kaelbling, L.~P., and Littman, M.~L.
\newblock Acting optimally in partially observable stochastic domains.
\newblock In \emph{Proceedings of the 12th National Conference on Artificial
  Intelligence, Seattle, WA, USA, July 31 - August 4, 1994, Volume 2}, pp.\
  1023--1028, 1994.

\bibitem[Espeholt et~al.(2018)Espeholt, Soyer, Munos, Simonyan, Mnih, Ward,
  Doron, Firoiu, Harley, Dunning, Legg, and Kavukcuoglu]{espeholt2018scalable}
Espeholt, L., Soyer, H., Munos, R., Simonyan, K., Mnih, V., Ward, T., Doron,
  Y., Firoiu, V., Harley, T., Dunning, I., Legg, S., and Kavukcuoglu, K.
\newblock {IMPALA:} scalable distributed deep-{RL} with importance weighted
  actor-learner architectures.
\newblock In \emph{Proceedings of the 35th International Conference on Machine
  Learning, {ICML} 2018, Stockholmsm{\"{a}}ssan, Stockholm, Sweden, July 10-15,
  2018}, pp.\  1406--1415, 2018.

\bibitem[Gregor et~al.(2015)Gregor, Danihelka, Graves, Rezende, and
  Wierstra]{gregor2015draw}
Gregor, K., Danihelka, I., Graves, A., Rezende, D.~J., and Wierstra, D.
\newblock {DRAW:} {A} recurrent neural network for image generation.
\newblock In \emph{Proceedings of the 32nd International Conference on Machine
  Learning, {ICML} 2015, Lille, France, 6-11 July 2015}, pp.\  1462--1471,
  2015.

\bibitem[Gregor et~al.(2019)Gregor, Jimenez~Rezende, Besse, Wu, Merzic, and
  van~den Oord]{gregor2019shaping}
Gregor, K., Jimenez~Rezende, D., Besse, F., Wu, Y., Merzic, H., and van~den
  Oord, A.
\newblock Shaping belief states with generative environment models for {RL}.
\newblock In Wallach, H., Larochelle, H., Beygelzimer, A., d\textquotesingle
  Alch\'{e}-Buc, F., Fox, E., and Garnett, R. (eds.), \emph{Advances in Neural
  Information Processing Systems 32}, pp.\  13475--13487. Curran Associates,
  Inc., 2019.

\bibitem[Guo et~al.(2018)Guo, Azar, Piot, Pires, Pohlen, and
  Munos]{guo2018neural}
Guo, Z.~D., Azar, M.~G., Piot, B., Pires, B.~A., Pohlen, T., and Munos, R.
\newblock Neural predictive belief representations.
\newblock \emph{arXiv preprint arXiv:1811.06407}, 2018.

\bibitem[Ha \& Schmidhuber(2018)Ha and Schmidhuber]{ha2018recurrent}
Ha, D. and Schmidhuber, J.
\newblock Recurrent world models facilitate policy evolution.
\newblock In \emph{Advances in Neural Information Processing Systems 31: Annual
  Conference on Neural Information Processing Systems 2018, NeurIPS 2018, 3-8
  December 2018, Montr{\'{e}}al, Canada}, pp.\  2455--2467, 2018.

\bibitem[Hafner et~al.(2019)Hafner, Lillicrap, Fischer, Villegas, Ha, Lee, and
  Davidson]{hafner2019learning}
Hafner, D., Lillicrap, T., Fischer, I., Villegas, R., Ha, D., Lee, H., and
  Davidson, J.
\newblock Learning latent dynamics for planning from pixels.
\newblock In Chaudhuri, K. and Salakhutdinov, R. (eds.), \emph{Proceedings of
  the 36th International Conference on Machine Learning}, volume~97 of
  \emph{Proceedings of Machine Learning Research}, pp.\  2555--2565, Long
  Beach, California, USA, 09--15 Jun 2019. PMLR.

\bibitem[He et~al.(2016{\natexlab{a}})He, Zhang, Ren, and Sun]{he2016deep}
He, K., Zhang, X., Ren, S., and Sun, J.
\newblock Deep residual learning for image recognition.
\newblock In \emph{2016 {IEEE} Conference on Computer Vision and Pattern
  Recognition, {CVPR} 2016, Las Vegas, NV, USA, June 27-30, 2016}, pp.\
  770--778, 2016{\natexlab{a}}.
\newblock \doi{10.1109/CVPR.2016.90}.

\bibitem[He et~al.(2016{\natexlab{b}})He, Zhang, Ren, and Sun]{he2016identity}
He, K., Zhang, X., Ren, S., and Sun, J.
\newblock Identity mappings in deep residual networks.
\newblock In \emph{European conference on computer vision}, pp.\  630--645.
  Springer, 2016{\natexlab{b}}.

\bibitem[Hessel et~al.(2019)Hessel, Soyer, Espeholt, Czarnecki, Schmitt, and
  van Hasselt]{hessel2019multi}
Hessel, M., Soyer, H., Espeholt, L., Czarnecki, W., Schmitt, S., and van
  Hasselt, H.
\newblock Multi-task deep reinforcement learning with {PopArt}.
\newblock In \emph{Proceedings of the AAAI Conference on Artificial
  Intelligence}, volume~33, pp.\  3796--3803, 2019.

\bibitem[Hochreiter \& Schmidhuber(1997)Hochreiter and
  Schmidhuber]{hochreiter1997long}
Hochreiter, S. and Schmidhuber, J.
\newblock Long short-term memory.
\newblock \emph{Neural Computation}, 9\penalty0 (8):\penalty0 1735--1780, 1997.
\newblock \doi{10.1162/neco.1997.9.8.1735}.

\bibitem[Houthooft et~al.(2016)Houthooft, Chen, Duan, Schulman, Turck, and
  Abbeel]{houthooft2016vime}
Houthooft, R., Chen, X., Duan, Y., Schulman, J., Turck, F.~D., and Abbeel, P.
\newblock {VIME:} variational information maximizing exploration.
\newblock In \emph{Advances in Neural Information Processing Systems 29: Annual
  Conference on Neural Information Processing Systems 2016, December 5-10,
  2016, Barcelona, Spain}, pp.\  1109--1117, 2016.

\bibitem[Jaderberg et~al.(2017)Jaderberg, Mnih, Czarnecki, Schaul, Leibo,
  Silver, and Kavukcuoglu]{jaderberg2017reinforcement}
Jaderberg, M., Mnih, V., Czarnecki, W.~M., Schaul, T., Leibo, J.~Z., Silver,
  D., and Kavukcuoglu, K.
\newblock Reinforcement learning with unsupervised auxiliary tasks.
\newblock In \emph{5th International Conference on Learning Representations,
  {ICLR} 2017, Toulon, France, April 24-26, 2017, Conference Track
  Proceedings}, 2017.

\bibitem[Kingma \& Welling(2014)Kingma and Welling]{kingma2013auto}
Kingma, D.~P. and Welling, M.
\newblock Auto-encoding variational bayes.
\newblock In \emph{2nd International Conference on Learning Representations,
  {ICLR} 2014, Banff, AB, Canada, April 14-16, 2014, Conference Track
  Proceedings}, 2014.

\bibitem[Levine et~al.(2016)Levine, Finn, Darrell, and Abbeel]{levine2016end}
Levine, S., Finn, C., Darrell, T., and Abbeel, P.
\newblock End-to-end training of deep visuomotor policies.
\newblock \emph{Journal of Machine Learning Research}, 17\penalty0
  (1):\penalty0 1334–1373, January 2016.
\newblock ISSN 1532-4435.

\bibitem[Littman et~al.(2001)Littman, Sutton, and Singh]{littman2002predictive}
Littman, M.~L., Sutton, R.~S., and Singh, S.~P.
\newblock Predictive representations of state.
\newblock In \emph{Advances in Neural Information Processing Systems 14 [Neural
  Information Processing Systems: Natural and Synthetic, {NIPS} 2001, December
  3-8, 2001, Vancouver, British Columbia, Canada]}, pp.\  1555--1561, 2001.

\bibitem[Mirowski et~al.(2017)Mirowski, Pascanu, Viola, Soyer, Ballard, Banino,
  Denil, Goroshin, Sifre, Kavukcuoglu, Kumaran, and
  Hadsell]{mirowski2016learning}
Mirowski, P., Pascanu, R., Viola, F., Soyer, H., Ballard, A., Banino, A.,
  Denil, M., Goroshin, R., Sifre, L., Kavukcuoglu, K., Kumaran, D., and
  Hadsell, R.
\newblock Learning to navigate in complex environments.
\newblock In \emph{5th International Conference on Learning Representations,
  {ICLR} 2017, Toulon, France, April 24-26, 2017, Conference Track
  Proceedings}, 2017.

\bibitem[Mnih et~al.(2015)Mnih, Kavukcuoglu, Silver, Rusu, Veness, Bellemare,
  Graves, Riedmiller, Fidjeland, Ostrovski, et~al.]{mnih2015human}
Mnih, V., Kavukcuoglu, K., Silver, D., Rusu, A.~A., Veness, J., Bellemare,
  M.~G., Graves, A., Riedmiller, M., Fidjeland, A.~K., Ostrovski, G., et~al.
\newblock Human-level control through deep reinforcement learning.
\newblock \emph{Nature}, 518\penalty0 (7540):\penalty0 529, 2015.

\bibitem[Mnih et~al.(2016)Mnih, Badia, Mirza, Graves, Lillicrap, Harley,
  Silver, and Kavukcuoglu]{mnih2016asynchronous}
Mnih, V., Badia, A.~P., Mirza, M., Graves, A., Lillicrap, T.~P., Harley, T.,
  Silver, D., and Kavukcuoglu, K.
\newblock Asynchronous methods for deep reinforcement learning.
\newblock In \emph{Proceedings of the 33nd International Conference on Machine
  Learning, {ICML} 2016, New York City, NY, USA, June 19-24, 2016}, pp.\
  1928--1937, 2016.

\bibitem[Moreno et~al.(2018)Moreno, Humplik, Papamakarios, Pires, Buesing,
  Heess, and Weber]{moreno2018neural}
Moreno, P., Humplik, J., Papamakarios, G., Pires, B.~A., Buesing, L., Heess,
  N., and Weber, T.
\newblock Neural belief states for partially observed domains.
\newblock In \emph{NeurIPS 2018 workshop on Reinforcement Learning under
  Partial Observability}, 2018.

\bibitem[Nair \& Hinton(2010)Nair and Hinton]{nair2010rectified}
Nair, V. and Hinton, G.~E.
\newblock Rectified linear units improve restricted boltzmann machines.
\newblock In \emph{Proceedings of the 27th International Conference on Machine
  Learning (ICML-10), June 21-24, 2010, Haifa, Israel}, pp.\  807--814, 2010.

\bibitem[Oh et~al.(2015)Oh, Guo, Lee, Lewis, and Singh]{oh2015action}
Oh, J., Guo, X., Lee, H., Lewis, R.~L., and Singh, S.~P.
\newblock Action-conditional video prediction using deep networks in {Atari}
  games.
\newblock In \emph{Advances in Neural Information Processing Systems 28: Annual
  Conference on Neural Information Processing Systems 2015, December 7-12,
  2015, Montreal, Quebec, Canada}, pp.\  2863--2871, 2015.

\bibitem[Oh et~al.(2017)Oh, Singh, and Lee]{oh2017value}
Oh, J., Singh, S., and Lee, H.
\newblock Value prediction network.
\newblock In \emph{Advances in Neural Information Processing Systems 30: Annual
  Conference on Neural Information Processing Systems 2017, 4-9 December 2017,
  Long Beach, CA, {USA}}, pp.\  6118--6128, 2017.

\bibitem[Oord et~al.(2018)Oord, Li, and Vinyals]{oord2018representation}
Oord, A. v.~d., Li, Y., and Vinyals, O.
\newblock Representation learning with contrastive predictive coding.
\newblock \emph{arXiv preprint arXiv:1807.03748}, 2018.

\bibitem[Pathak et~al.(2017)Pathak, Agrawal, Efros, and
  Darrell]{pathak2017curiosity}
Pathak, D., Agrawal, P., Efros, A.~A., and Darrell, T.
\newblock Curiosity-driven exploration by self-supervised prediction.
\newblock In \emph{Proceedings of the 34th International Conference on Machine
  Learning, {ICML} 2017, Sydney, NSW, Australia, 6-11 August 2017}, pp.\
  2778--2787, 2017.

\bibitem[Puigdom{\`e}nech~Badia et~al.(2020)Puigdom{\`e}nech~Badia, Sprechmann,
  Vitvitskyi, Guo, Piot, Kapturowski, Tieleman, Arjovsky, Pritzel, Bolt, and
  Blundell]{badia2020never}
Puigdom{\`e}nech~Badia, A., Sprechmann, P., Vitvitskyi, A., Guo, D., Piot, B.,
  Kapturowski, S., Tieleman, O., Arjovsky, M., Pritzel, A., Bolt, A., and
  Blundell, C.
\newblock Never give up: Learning directed exploration strategies.
\newblock In \emph{International Conference on Learning Representations}, 2020.

\bibitem[Schrittwieser et~al.(2019)Schrittwieser, Antonoglou, Hubert, Simonyan,
  Sifre, Schmitt, Guez, Lockhart, Hassabis, Graepel,
  et~al.]{schrittwieser2019mastering}
Schrittwieser, J., Antonoglou, I., Hubert, T., Simonyan, K., Sifre, L.,
  Schmitt, S., Guez, A., Lockhart, E., Hassabis, D., Graepel, T., et~al.
\newblock Mastering atari, go, chess and shogi by planning with a learned
  model.
\newblock \emph{arXiv preprint arXiv:1911.08265}, 2019.

\bibitem[Silver et~al.(2017)Silver, Schrittwieser, Simonyan, Antonoglou, Huang,
  Guez, Hubert, Baker, Lai, Bolton, et~al.]{silver2017mastering}
Silver, D., Schrittwieser, J., Simonyan, K., Antonoglou, I., Huang, A., Guez,
  A., Hubert, T., Baker, L., Lai, M., Bolton, A., et~al.
\newblock Mastering the game of {Go} without human knowledge.
\newblock \emph{nature}, 550\penalty0 (7676):\penalty0 354--359, 2017.

\bibitem[Singh et~al.(2003)Singh, Littman, Jong, Pardoe, and
  Stone]{singh2003learning}
Singh, S.~P., Littman, M.~L., Jong, N.~K., Pardoe, D., and Stone, P.
\newblock Learning predictive state representations.
\newblock In \emph{Machine Learning, Proceedings of the Twentieth International
  Conference {(ICML} 2003), August 21-24, 2003, Washington, DC, {USA}}, pp.\
  712--719, 2003.

\bibitem[Song et~al.(2020)Song, Abdolmaleki, Springenberg, Clark, Soyer, Rae,
  Noury, Ahuja, Liu, Tirumala, Heess, Belov, Riedmiller, and
  Botvinick]{song2019vmpo}
Song, H.~F., Abdolmaleki, A., Springenberg, J.~T., Clark, A., Soyer, H., Rae,
  J.~W., Noury, S., Ahuja, A., Liu, S., Tirumala, D., Heess, N., Belov, D.,
  Riedmiller, M., and Botvinick, M.~M.
\newblock {V-MPO:} {O}n-policy maximum a posteriori policy optimization for
  discrete and continuous control.
\newblock In \emph{(to appear) 8th International Conference on Learning
  Representations, {ICLR} 2020}, 2020.

\bibitem[Sutton \& Barto(2018)Sutton and Barto]{sutton2018reinforcement}
Sutton, R.~S. and Barto, A.~G.
\newblock \emph{Reinforcement learning: An introduction}.
\newblock MIT press, 2018.

\bibitem[Szepesv{\'a}ri(2010)]{szepesvari2010algorithms}
Szepesv{\'a}ri, C.
\newblock Algorithms for reinforcement learning.
\newblock \emph{Synthesis lectures on artificial intelligence and machine
  learning}, 4\penalty0 (1):\penalty0 1--103, 2010.

\bibitem[Tassa et~al.(2018)Tassa, Doron, Muldal, Erez, Li, Casas, Budden,
  Abdolmaleki, Merel, Lefrancq, et~al.]{tassa2018deepmind}
Tassa, Y., Doron, Y., Muldal, A., Erez, T., Li, Y., Casas, D. d.~L., Budden,
  D., Abdolmaleki, A., Merel, J., Lefrancq, A., et~al.
\newblock {DeepMind} control suite.
\newblock \emph{arXiv preprint arXiv:1801.00690}, 2018.

\bibitem[Vinyals et~al.(2019)Vinyals, Babuschkin, Czarnecki, Mathieu, Dudzik,
  Chung, Choi, Powell, Ewalds, Georgiev, et~al.]{vinyals2019grandmaster}
Vinyals, O., Babuschkin, I., Czarnecki, W.~M., Mathieu, M., Dudzik, A., Chung,
  J., Choi, D.~H., Powell, R., Ewalds, T., Georgiev, P., et~al.
\newblock Grandmaster level in {StarCraft II} using multi-agent reinforcement
  learning.
\newblock \emph{Nature}, 575\penalty0 (7782):\penalty0 350--354, 2019.

\bibitem[Zhang et~al.(2019)Zhang, Vikram, Smith, Abbeel, Johnson, and
  Levine]{zhang2018solar}
Zhang, M., Vikram, S., Smith, L., Abbeel, P., Johnson, M.~J., and Levine, S.
\newblock {SOLAR:} {D}eep structured representations for model-based
  reinforcement learning.
\newblock In \emph{Proceedings of the 36th International Conference on Machine
  Learning, {ICML} 2019, 9-15 June 2019, Long Beach, California, {USA}}, pp.\
  7444--7453, 2019.

\end{thebibliography}

\clearpage
\newpage
\appendix
\onecolumn

\section{Implementation Details}
\label{sec:implementation}

In this section, we describe our implementation in more detail, especially architecture and parameter choices.
For hyperparameters picked from sweeps, we describe our protocol and the values considered.
We also indicate software libraries used and describe our protocol for plotting.

\subsection{Agent Architecture}

\Cref{tab:mainArchitecture} collects the different architecture parameters for the ``main agent networks''. These include the observation processing, the RNN for full histories, and the MLPs for value estimates and policies.
Our choices follow \citet{hessel2019multi} with a few exceptions that are given in bold in \cref{tab:mainArchitecture}.
The differences are limited to increases in network size.
The RNNs used are LSTMS \citep{hochreiter1997long} and the networks used for image processing are ResNets \citep{he2016deep}
\begin{table}[htb]
    \centering
    \begin{tabular}{lc}
        Hyperparameter & Value \\
        \toprule
        Convolutional Stack &   \\
        \cmidrule(l){1-1}
        Number of sections & \texttt{3} \\
        Channels per section &  \texttt{(16, 32, 32)} \\
        Activation function &  \texttt{ReLU} \\
        \midrule
        ResNet section &   \\
        \cmidrule(l){1-1}
        Number of sections & \texttt{1 / 3x3 / 1} \\
        Max-Pool & \texttt{1 / 3x3 / 2} \\
        Conv & \texttt{2 / 3x3 / 1} \\
        Skip & \texttt{Identity} \\
        Conv & \texttt{2 / 3x3 / 1} \\
        Skip & \texttt{Identity} \\
        \midrule
        Post-ResNet &   \\
        \cmidrule(l){1-1}
        Fully connected layer & \textbf{\texttt{512}} \\
        Output activation &  \texttt{ReLU} \\
        \midrule
        Language preprocessing &   \\
        \cmidrule(l){1-1}
        Word embeddings & \texttt{20} \\
        Sentence embedding & \texttt{LSTM 64} \\ 
        Vocabulary size & \texttt{1000} \\
        Max.~instruction length & \texttt{15} \\
        \midrule
        Other input preprocessing &   \\
        \cmidrule(l){1-1}
        Action & one-hot \\
        Reward & see \cref{eq:rewardTransform} \\
        \midrule
        Full history compression & \\
        \cmidrule(l){1-1}
        $h_{\mathrm{f}}$ & \texttt{LSTM \textbf{(512, 512)}} \\
        Skip connections & \texttt{\textbf{Yes}} \\
        \midrule
        RL heads & \\
        \cmidrule(l){1-1}
        Hidden layer activation &  \texttt{ReLU} \\
        Value & \textbf{\texttt{(512, 30)}} \\    
        Policy (softmax) & \textbf{\texttt{(512, 15)}} \\
        \bottomrule
    \end{tabular}
    \caption{Main architecture parameters. 
    These are the same as Table 7 of \citet{hessel2019multi}, except for parameters in bold.}
    \label{tab:mainArchitecture}
\end{table}
The DMLab30 observations we use and the way we process them is the same as \citet{hessel2019multi} (with the exception of the size of the post-resnet layer).
The agent's first-person view is processed through a ResNet \citep{he2016identity}, which is then flattened and input to a fully connected layer, with ReLU \citep{nair2010rectified} output activations.
The language instructions are processed as follows.
Each word is embedded into a $20$-dimensional vector, where the vocabulary size is $100$ and the embedding is learnable.
The first $15$ embedded words (or less if the instruction is shorter) are fed into an LSTM with $64$ hidden units.
The outputs are summed into the final instruction embedding.
The previous action (which resulted in the given first-person view) is encoded as a one-hot, out of the $15$ possible actions (from the action set given in table 7 of \citet{hessel2019multi}).
The reward observed along with the first-person view (resulting from the ``previous action'') is transformed according to
\begin{equation}
r \mapsto 0.3 \cdot \tanh(r/5)_- + 1.5 \cdot \tanh(r/5)_+,
    \label{eq:rewardTransform}
\end{equation}
where $(\cdot)_-$ denotes the negative part and $(\cdot)_+$ the positive part.
The processed first-person view, instruction, previous action and reward are concatenated and fed as input to the full history RNN $h_{\mathrm{f}}$.
The output of the full history RNN is input to the value estimate MLP, which has 30 outputs because PopArt maintains one value estimate per task.
Note that PopArt uses the task index for updating these estimates \citet{hessel2019multi}.
This task index is not an input to any network.
The output of the full history RNN is also input to the policy head, which takes a softmax over $15$ logits corresponding to the actions in the discretized action set introduced by \citet{hessel2019multi}.

\subsection{Action-conditional predictions of the future}
\label{app:implementation:acpf}

The procedure for compressing partial histories is the same for \acronym{}, CPC and Simcore DRAW, and is detailed in \cref{tab:futurePredictions}.
\begin{table}[htb]
    \centering
    \begin{tabular}{lc}
        Hyperparameter & Value \\
        \toprule
        $h_{\mathrm{p}}$ & \texttt{LSTM (512, 512)} \\
        Future prediction horizon & \texttt{20} \\
        Time subsampling & to size \texttt{6} \\
        Batch subsampling & None \\
        Future subsampling & to size \texttt{2} \\
        \bottomrule
    \end{tabular}
    \caption{Parameters for action-conditional predictions of the future.}
    \label{tab:futurePredictions}
\end{table}
Our implementation reproduces the Simcore of \citet{gregor2019shaping}.

The compressed full histories are built by unrolling $h_{\mathrm{f}}$ (see \cref{tab:mainArchitecture}).
Specifically, when $B^\theta_t$ is used as an input to MLPs, we use the LSTM output.
When $B^\theta_t$ is used as an input to an RNN (for unrolling $h_{\mathrm{f}}$ and $h_{\mathrm{p}}$), we use the LSTM state.
The LSTM for partial histories has the same architecture as the full history LSTM, and hence the same state size.
So we can generate $B^\theta_{t,k}$ by unrolling $h_{\mathrm{p}}$ with the action sequence $A_{t},\ldots,A_{t+k-1}$ (one-hot encoded), from the initial state provided by $h_{\mathrm{f}}$, namely the state resulting from the RNN update $h_{\mathrm{f}}(B_t, O_t, A_{t-1})$.

Our minibatches have size $\mathtt{T} \times \mathtt{B}$, where $\mathtt{T}$ is the sequence length and $\mathtt{B}$ is the batch size (see \cref{tab:rlTraining}).
In addition to the sequences of length $\mathtt{T}$, we have the $\mathtt{B}$ LSTM states of the main architecture at the start of these sequences \citep{espeholt2018scalable}.

We do not use the initial states as compressed full histories, so there are $\mathtt{T} \times \mathtt{B}$ possible compressed full histories to be used.
Before we unroll $h_{\mathrm{p}}$, we subsample the time indices and use only $\mathtt{6} \times \mathtt{B}$ compressed full histories (the randomly selected time indices are the same across the batch axis, but change in each minibatch).
From these $\mathtt{6} \times \mathtt{B}$ initial states, we generate $\mathtt{H} \times \mathtt{6} \times \mathtt{B}$ partial histories.
We used $\mathtt{H} = 20$,
except in the scalability experiments, where we varied $\mathtt{H}$ across $(\mathtt{1}, \mathtt{5}, \mathtt{20})$.
We then subsampled these partial histories, taking $\mathtt{2} \times \mathtt{6} \times \mathtt{B}$ to be used for representation learning (\acronym{}, CPC or DRAW).
The randomly selected future indices are the same across the batch and time axis, but change in each minibatch.

\subsection{Picking Hyperparameters for Representation Learning}

The weight of the representation learning losses for all methods (\acronym{}, Simcore DRAW, CPC and pixel control) were chosen for top performance for each method, chosen among powers of $10$ from $-2$ to $2$.

The prediction horizon (how far we predict into the future) for \acronym{}, Simcore DRAW, CPC has been fixed at $20$.
We have tried changing this horizon to $1$ and to $5$, for all methods, similar to the study for \acronym{} presented in \cref{sec:experiments}.
With the exception of \acronym{}, predicting up to twenty steps into the future did not yield improvements compared to only up to five steps, whereas predicting up to five steps into the future was marginally better (for Simcore DRAW) or the same (for CPC) than using a horizon of one step.
In all cases, therefore, $20$ was a best performing choice for the horizon.

\subsection{\acronym{}}

The representation learning in \acronym{} is setup as follows.
The forward prediction, from partial histories to future latents, uses the procedure described in \cref{app:implementation:acpf}.
The reverse prediction is from latents to full histories  without subsampling, over the whole $\mathtt{T} \times \mathtt{B}$ minibatch.

\Cref{tab:pbl} gives the breakdown of parameters for \acronym{}.
\begin{table}[htb]
    \centering
    \begin{tabular}{lc}
        Hyperparameter & Value \\
        \toprule
        Latent Embedding & \\
        \cmidrule(l){1-1}
        $f$ & same as in \cref{tab:mainArchitecture} \\
        Shared $f$ & \texttt{No} \\
        \midrule
        Forward prediction &   \\
        \cmidrule(l){1-1}
        Loss weight & \texttt{1} \\
        $g$ & \texttt{(512, 512, 592)} \\
        Normalize latents (targets) & \texttt{Yes} \\
        Normalize predictions & \texttt{Yes} \\
        Prediction regularizer & $v \mapsto 0.02 \cdot (\|v\|^2_2 - 1)^2$ \\
        \midrule
        Reverse prediction &   \\
        \cmidrule(l){1-1}
        Loss weight & \texttt{1} \\
        $g'$ & \texttt{(512, 512, 1024)} \\
        Normalize latents ($g'$ inputs) & \texttt{Yes} \\
        Normalize predictions & \texttt{No} \\
        Prediction regularizer & None \\
        Latent regularizer & $v \mapsto 0.02 \cdot (\|v\|^2_2 - 1)^2$ \\
        \midrule
        All hidden layer activations &  \texttt{ReLU} \\
        \bottomrule
    \end{tabular}
    \caption{\acronym{} parameters.}
    \label{tab:pbl}
\end{table}
We use the architecture detailed in \cref{tab:mainArchitecture} to process observations ($f$), but in a separate network from the one used by the agent architecture.
The output of $f$ is a vector of size $592$ ($512$ for first-person view, $64$ for language, $15$ for the previous action and $1$ for the reward), which is the latent embedding of the observation.
The output of $g$ is a vector of size $592$, the output size of $f$.
The output of $g'$ is a vector of size $1024$, the output size of $h_{\mathrm{f}}$ (which has two layers of size $512$ with skip connections, as given in \cref{tab:mainArchitecture}).

The latent embeddings are always normalized with $v \mapsto \frac{v}{\|v\|_2 + 10^{-8}}$, both when used as targets in the forward prediction and as inputs to $g'$ in the reverse prediction.
For the forward prediction, the predictions (outputs of $g$) are also normalized in this way, and regularized to have to have unit $\ell_2$ norm (the regularization is applied before normalization).
In the reverse prediction, the embeddings are regularized (outputs of $f$) to have unit $\ell_2$ norm (again, the regularization is applied before normalization).
The compressed histories (outputs of $h_{\mathrm{f}}$ and $h_{\mathrm{p}}$) and predictions (outputs of $g'$) are neither normalized nor regularized.

\subsection{CPC}
\label{app:CPC}

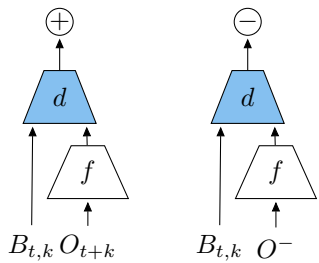
\begin{figure}[htb]
    \centering
    \begin{tikzpicture}[shorten >=1pt,->, node distance=\layersep]
        \tikzstyle{every pin edge}=[<-,shorten <=1pt]
        \tikzstyle{mlp}=[shape=trapezium,draw=black,fill=black!25,minimum size=2em,trapezium stretches body]
    
        \node[shape=circle,draw=black,fill=white,inner sep=0pt] (l_p) at (-1.25, 1) {$+$};
        \node[mlp,fill=zesty_3] (d_p) at (-1.25, 0) {$d$};
        \node[mlp,fill=white] (f_p) at (-0.875, -1) {$f$};
        \node[] (b_p) at (-1.625, -2) {$B_{t,k}$};
        \node[] (o_p) at (-0.875, -2) {$O_{t+k}$};
        
        \draw [-gradTip] (o_p) -- (f_p);
        \draw [-gradTip] (f_p) -- (d_p.south east);
        \draw [-gradTip] (b_p) -- (d_p.south west);
        \draw [-gradTip] (d_p) -- (l_p);
        
        \node[shape=circle,draw=black,fill=white,inner sep=0pt] (l_n) at (1.25, 1) {$-$};
        \node[mlp,fill=zesty_3, trapezium stretches body] (d_n) at (1.25, 0) {$d$};
        \node[mlp,fill=white] (f_n) at (1.625, -1) {$f$};
        \node[] (b_n) at (0.875, -2) {$B_{t,k}$};
        \node[] (o_n) at (1.625, -2) {$O^-$};
        
        \draw [-gradTip] (o_n) -- (f_n);
        \draw [-gradTip] (f_n) -- (d_n.south east);
        \draw [-gradTip] (b_n) -- (d_n.south west);
        \draw [-gradTip] (d_n) -- (l_n);
    \end{tikzpicture}
    \caption{Architecture diagram for contrastive predictive coding.}
    \label{fig:CPC}
\end{figure}

CPC aims to learn a discriminator between jointly distributed random variables and random variables sampled from the marginals.
A diagram of CPC is given in \cref{fig:CPC}, illustrating discrimination between jointly distributed partial histories and observations, from independent ones. 
The parameters are given in \cref{tab:cpc}.
To sample negative examples for timestep $t+k$, we take twenty observations at random, with replacement, from the minibatch, uniformly over the indices different from $t+k$.
The losses over positive and negative examples are balanced---the total loss is the binary (sigmoid) cross-entropy over the positive example plus the mean over all negative examples of the binary (sigmoid) cross-entropy.
\begin{table}[htb]
    \centering
    \begin{tabular}{lc}
        Hyperparameter & Value \\
        \toprule
        Loss weight & \texttt{0.1} \\
        Negative examples & \texttt{20} \\
        $d$ & \texttt{(512, 512, 1)} \\
        $f$ & see \cref{tab:mainArchitecture} \\
        Shared $f$ & \texttt{Yes} \\
        \bottomrule
    \end{tabular}
    \caption{CPC parameters.}
    \label{tab:cpc}
\end{table}

\subsection{DRAW}

\Cref{tab:draw} outlines parameters for DRAW and GECO. 
Parameters not reported in \cref{tab:mainArchitecture,tab:futurePredictions,tab:draw} are the same as reported by \citet{gregor2019shaping}.
\begin{table}[htb]
    \centering
    \begin{tabular}{lc}
        Hyperparameter & Value \\
        \toprule
        Loss weight & \texttt{\textbf{1e-1}} \\
        GECO $\kappa$ & \texttt{0.015} \\
        \bottomrule
    \end{tabular}
    \caption{DRAW loss weight. 
    See \citet{gregor2019shaping} for full details, and \cref{tab:mainArchitecture,tab:futurePredictions}.}
    \label{tab:draw}
\end{table}

\subsection{Pixel Control}

\Cref{tab:pc} shows the parameters for pixel control.
Parameters not reported as as used by \citet{hessel2019multi} (see section ``Pixel Control'' in their appendix). 
Differently from \citet{hessel2019multi}, we compute the total loss over $4 \times 4$ cells by taking the mean loss instead of the sum of losses.
\begin{table}[htb]
    \centering
    \begin{tabular}{lc}
        Hyperparameter & Value \\
        \toprule
        Loss weight & \texttt{\textbf{1e-1}} \\
        Total loss & \textbf{mean} over cells \\
        \bottomrule
    \end{tabular}
    \caption{Pixel control parameters. 
    See \citet{hessel2019multi} for full details.
    Different parameters are designated in bold.}
    \label{tab:pc}
\end{table}

\subsection{VTrace and PopArt}

The RL and other training parameters are given in \cref{tab:rlTraining} and PopArt in \cref{tab:popart}.
We use the same parameters as \citet{hessel2019multi} except where designated in bold.
We did not use population-based training (PBT), so we tuned the entropy cost with a over $\{0.01, 0.005, 0.001, 0.0005\}$.
We chose the parameter that worked best for the base RL algorithm (IMPALA PopArt) with no representation learning, and used it for all the representation learning techniques.
\begin{table}[!h]
    \centering
    \begin{tabular}{lc}
        Hyperparameter & Value \\
        \toprule
        Statistics learning rate & \texttt{3e-4} \\
        Scale lower bound & \texttt{1e-4} \\
        Scale upper bound & \texttt{1e6} \\
        \bottomrule
    \end{tabular}
    \caption{PopArt parameters, the same as used by \citet{hessel2019multi}.}
    \label{tab:popart}
\end{table}
\begin{table}[!h]
    \centering
    \begin{tabular}{lc}
        Hyperparameter & Value \\
        \toprule
        RL & \\
        \cmidrule(l){1-1}
        Unroll length & \texttt{100} \\
        Batch size & \texttt{32} \\
        $\gamma$ & \texttt{0.99} \\
        V-trace $\lambda$ & \texttt{\textbf{0.99}} \\
        Baseline loss weight & \texttt{\textbf{0.4}} \\
        Entropy cost & \texttt{\textbf{5e-3}} \\
        \midrule
        DMLab 30 &  \\
        \cmidrule(l){1-1}
        First-person view (height, width) & \texttt{(72, 96)} \\
        Action repeats & \texttt{4} \\
        \midrule
        Optimizer & \texttt{\textbf{Adam}}  \\
        \cmidrule(l){1-1}
        Learning rate & \texttt{\textbf{1e-4}} \\
        $\beta_1$ & \texttt{\textbf{0}} \\
        $\beta_2$ & \texttt{\textbf{0.999}} \\
        $\epsilon$ & \texttt{\textbf{1e-6}} \\
        \bottomrule
    \end{tabular}
    \caption{RL and other training parameters.
    These match the parameters used by \citet{hessel2019multi} with differences in bold.}
    \label{tab:rlTraining}
\end{table}

\subsection{Libraries}

We used TensorFlow \citep{abadi2015tensorflow} and Sonnet (\url{github.com/deepmind/sonnet}).
TRFL (\url{github.com/deepmind/trfl}) provides the pixel control implementation we used.
In addition to implementations of convolutional layers, LSTMs and MLPs, Sonnet also provides the implementation we used for the language embedding.

\subsection{Plotting}
\label{app:plotting}

For our plots, we binned the observed per-episode returns into bins of size $10^8$ frames (centered at multiples of $10^8$) and took the mean performance in each bin.
These averaged returns were then mapped to the mean human normalized score, capped where applicable, and then averaged over tasks.
At this point we computed any confidence intervals if applicable, and finally averaged over independent runs.
For plots with confidence bands and \cref{tab:finalPerformance}, we assumed the mean estimates follow Student's $t$ distribution, and reported $95\%$ confidence intervals (two-sided).

The mean estimates in \cref{fig:breakdown,tab:finalPerformance} were taken by averaging the performances on the final $500\mathrm{M}$ frames of training.
The confidence intervals in \cref{fig:breakdown} were computed by adding the confidence intervals of the estimates for \acronym{} and pixel control.

\clearpage

\section{Supplementary results}
\label{sec:extra}

\subsection{DMLab-30}
\label{sec:DMLab30additional}
\begin{figure}[!h]
    \centering
    \begin{subfigure}{.45\textwidth}
        \includegraphics[width=.99\linewidth]{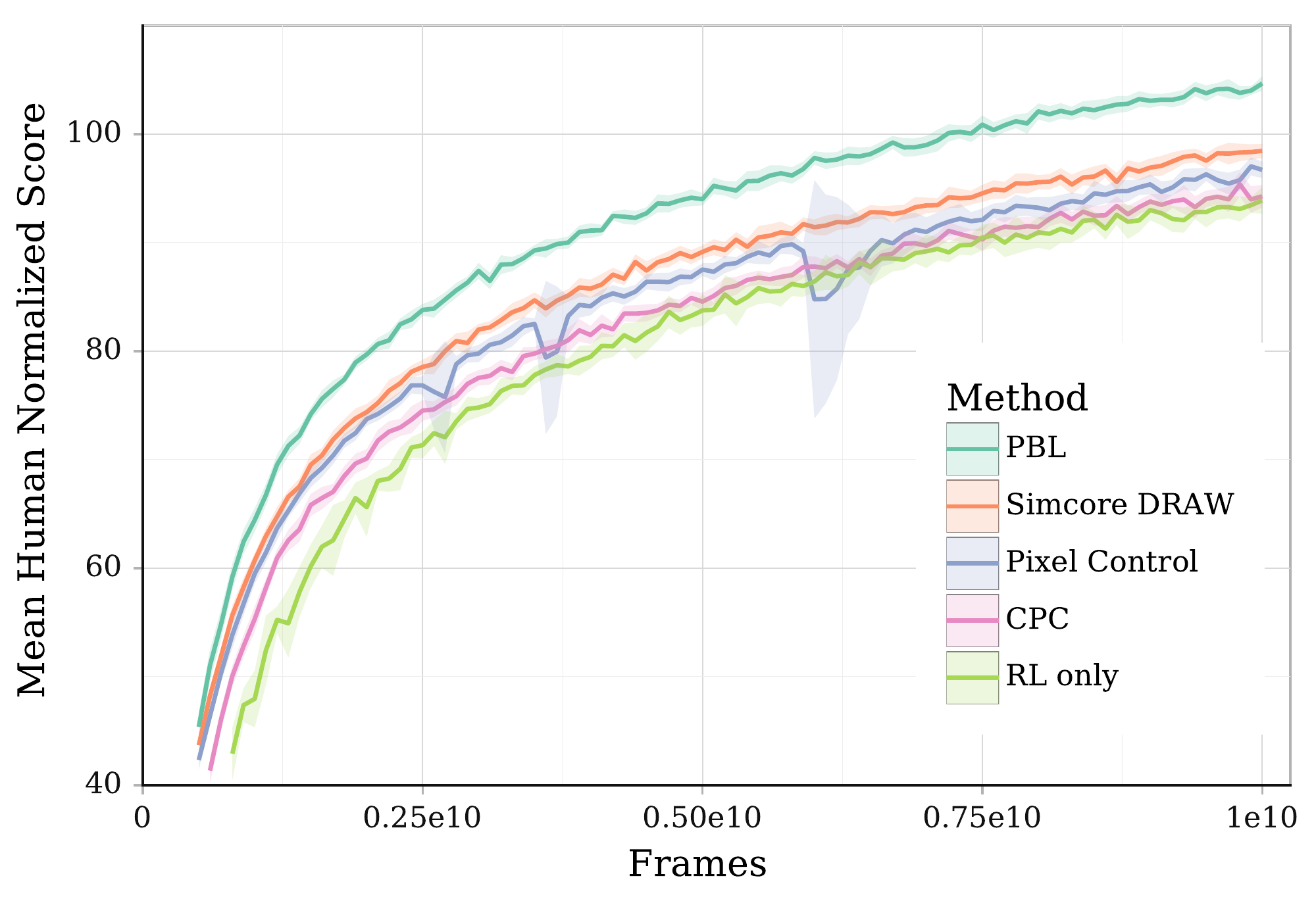}
        \caption{Mean uncapped human normalized score for compared methods.}
        \label{fig:comparison-uncapped}
    \end{subfigure}
    \quad
    \begin{subfigure}{.45\textwidth}
        \includegraphics[width=.99\linewidth]{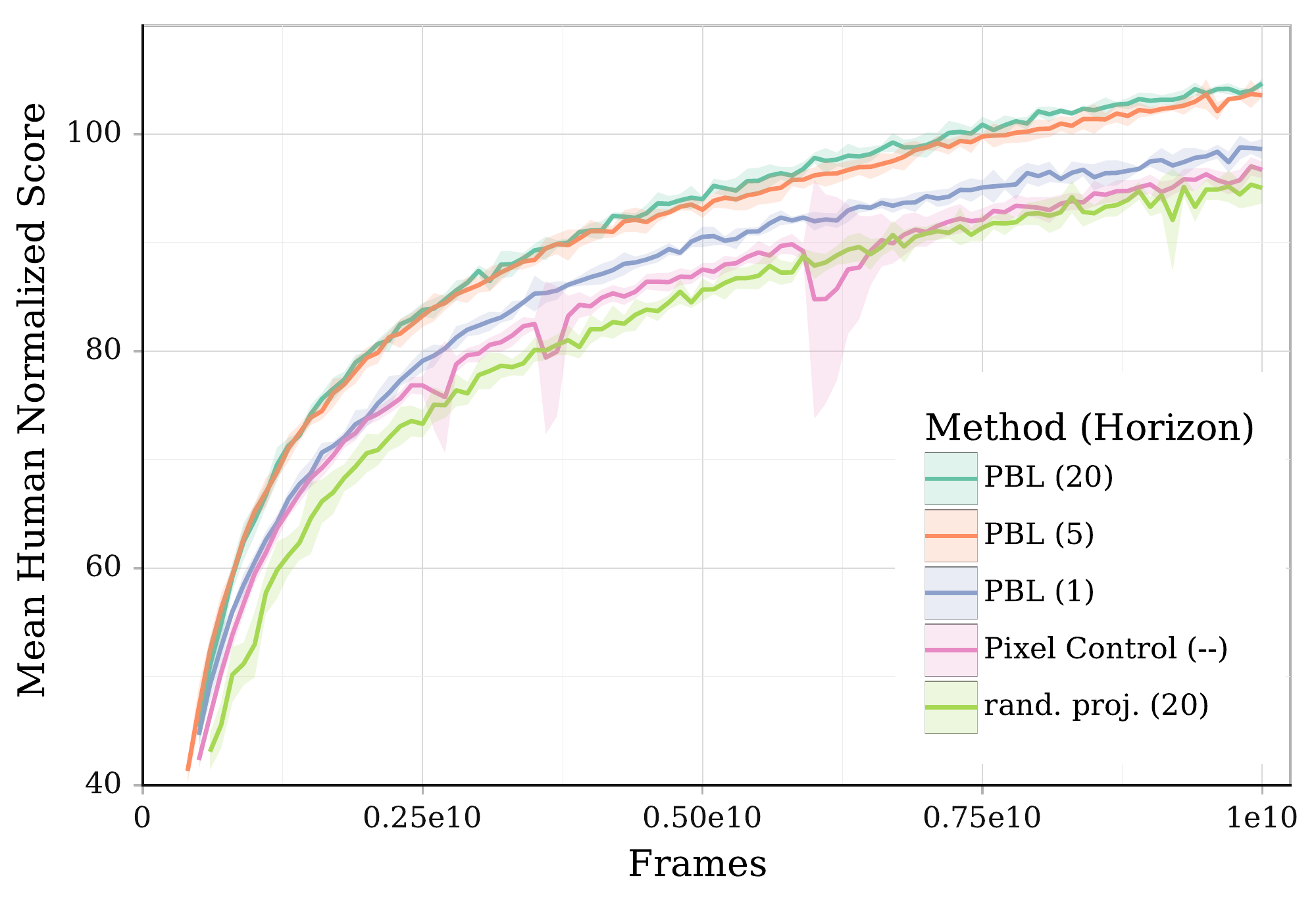}
        \caption{Mean uncapped human normalized score for \acronym{} across different prediction horizons.}
        \label{fig:scalability-uncapped}
    \end{subfigure}
\end{figure}

\begin{figure}[!h]
    \centering
    \includegraphics[width=8cm]{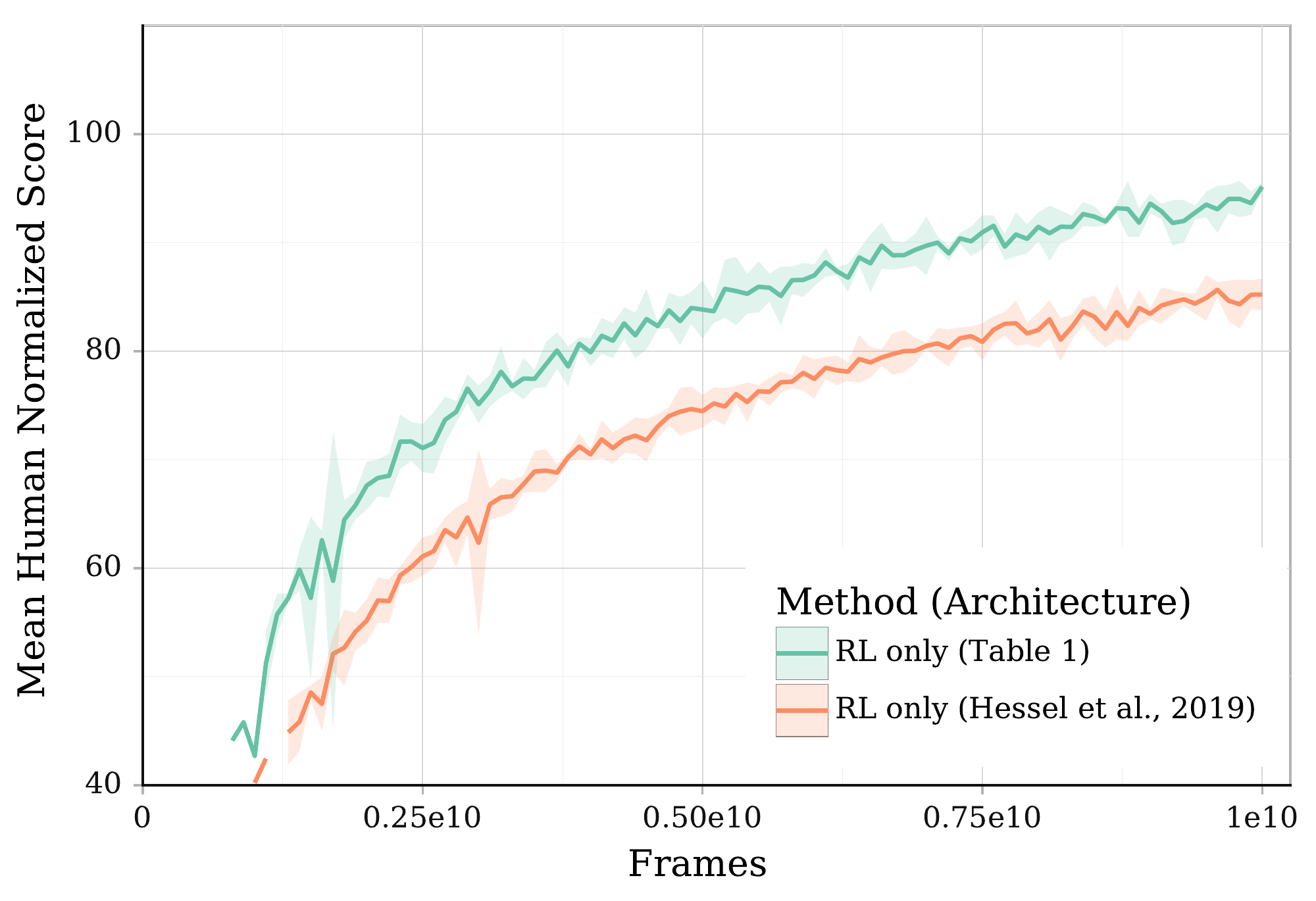}
    \caption{Mean uncapped human normalized score for compared architectures.}
    \label{fig:comparison-impala-uncapped}
\end{figure}

In this section, we present some additional results to provide more context to our results in DMLab-30.
\Cref{tab:finalPerformance} shows the final performance breakdown across all tasks in DMLab-30 for all compared methods. 
We also show the uncapped mean human normalized score for different methods (\cref{fig:comparison-uncapped}), for \acronym{} across different horizons (\cref{fig:scalability-uncapped}), and for the architecture comparison (\cref{fig:comparison-impala-uncapped}).

\begin{table*}
  \centering
  \small
  \begin{tabular}{lccccc}
  \toprule
     & \multicolumn{1}{c}{RL only} & \multicolumn{1}{c}{Pixel Control} & \multicolumn{1}{c}{CPC} & \multicolumn{1}{c}{SC DRAW} & \multicolumn{1}{c}{PBL} \\
    \midrule 
    {\small \bf \texttt{explore goal locations large}} & $\hphantom{0}77.0 \pm 3.4$ & $\hphantom{0}79.9 \pm 2.2$ & $\hphantom{0}76.9 \pm 1.3$ & $\hphantom{0}83.0 \pm 1.4$ & $\mathbf{\hphantom{0}90.7 \pm 3.0}$ \\
    {\small \bf \texttt{explore goal locations small}} & $123.6 \pm 3.2$ & $124.6 \pm 1.4$ & $124.5 \pm 2.1$ & $130.8 \pm 1.4$ & $\mathbf{137.7 \pm 1.3}$ \\
    {\small \bf \texttt{explore object locations large}} & $\hphantom{0}93.3 \pm 3.1$ & $\hphantom{0}97.5 \pm 1.5$ & $\hphantom{0}94.2 \pm 1.5$ & $102.1 \pm 1.3$ & $\mathbf{112.9 \pm 0.9}$ \\
    {\small \bf \texttt{explore object locations small}} & $106.0 \pm 2.6$ & $111.3 \pm 1.4$ & $108.7 \pm 1.2$ & $115.5 \pm 0.9$ & $\mathbf{124.8 \pm 1.0}$ \\
    {\small \bf \texttt{explore object rewards few}} & $\hphantom{0}57.9 \pm 1.1$ & $\hphantom{0}60.9 \pm 0.7$ & $\hphantom{0}59.8 \pm 0.7$ & $\hphantom{0}61.5 \pm 0.9$ & $\mathbf{\hphantom{0}64.9 \pm 0.5}$ \\
    {\small \bf \texttt{explore object rewards many}} & $\hphantom{0}59.7 \pm 0.7$ & $\hphantom{0}61.9 \pm 0.6$ & $\hphantom{0}61.3 \pm 0.7$ & $\hphantom{0}63.2 \pm 0.4$ & $\mathbf{\hphantom{0}67.7 \pm 0.5}$ \\
    {\small \texttt{explore obstructed goals large}} & $\hphantom{0}72.5 \pm 3.8$ & $\hphantom{0}79.9 \pm 2.5$ & $\hphantom{0}72.4 \pm 2.3$ & $\hphantom{0}78.9 \pm 2.1$ & $\hphantom{0}84.8 \pm 2.7$ \\
    {\small \bf \texttt{explore obstructed goals small}} & $114.8 \pm 3.7$ & $121.9 \pm 2.4$ & $118.7 \pm 2.3$ & $124.2 \pm 2.0$ & $\mathbf{133.3 \pm 1.2}$ \\
    {\small \texttt{language answer quantitative question}} & $159.5 \pm 6.1$ & $159.0 \pm 2.5$ & $158.9 \pm 2.6$ & $162.6 \pm 1.2$ & $165.0 \pm 1.4$ \\
    {\small \bf \texttt{language execute random task}} & $139.2 \pm 2.9$ & $141.2 \pm 3.1$ & $138.8 \pm 3.0$ & $145.4 \pm 2.1$ & $\mathbf{152.5 \pm 1.5}$ \\
    {\small \texttt{language select described object}} & $152.0 \pm 1.1$ & $154.5 \pm 1.9$ & $154.2 \pm 1.1$ & $158.2 \pm 0.7$ & $159.7 \pm 1.3$ \\
    {\small \texttt{language select located object}} & $241.8 \pm 6.9$ & $252.8 \pm 4.9$ & $243.9 \pm 5.6$ & $262.0 \pm 1.5$ & $263.9 \pm 1.1$ \\
    {\small \bf \texttt{lasertag one opponent large}} & $101.3 \pm 4.3$ & $112.4 \pm 4.6$ & $107.9 \pm 4.7$ & $111.7 \pm 2.2$ & $\mathbf{124.0 \pm 3.2}$ \\
    {\small \bf \texttt{lasertag one opponent small}} & $163.2 \pm 4.0$ & $168.5 \pm 3.5$ & $166.1 \pm 3.8$ & $171.2 \pm 2.6$ & $\mathbf{182.5 \pm 2.8}$ \\
    {\small \bf \texttt{lasertag three opponents large}} & $152.4 \pm 5.6$ & $158.1 \pm 2.7$ & $153.6 \pm 3.5$ & $158.0 \pm 1.7$ & $\mathbf{167.2 \pm 2.1}$ \\
    {\small \bf \texttt{lasertag three opponents small}} & $143.9 \pm 3.8$ & $148.9 \pm 1.9$ & $145.0 \pm 3.2$ & $147.4 \pm 1.7$ & $\mathbf{154.9 \pm 1.5}$ \\
    {\small \bf \texttt{natlab fixed large map}} & $\hphantom{0}41.7 \pm 3.1$ & $\hphantom{0}39.4 \pm 2.0$ & $\hphantom{0}38.9 \pm 2.2$ & $\hphantom{0}49.6 \pm 4.1$ & $\mathbf{\hphantom{0}93.3 \pm 7.9}$ \\
    {\small \texttt{natlab varying map randomized}} & $\hphantom{0}84.4 \pm 2.8$ & $\hphantom{0}85.2 \pm 2.8$ & $\hphantom{0}82.9 \pm 3.4$ & $\hphantom{0}82.5 \pm 2.9$ & $\hphantom{0}86.0 \pm 2.0$ \\
    {\small \bf \texttt{natlab varying map regrowth}} & $\hphantom{0}90.2 \pm 3.9$ & $\hphantom{0}87.6 \pm 4.3$ & $\hphantom{0}88.4 \pm 4.4$ & $\hphantom{0}92.4 \pm 2.1$ & $\mathbf{100.4 \pm 1.9}$ \\
    {\small \texttt{psychlab arbitrary visuomotor mapping}} & $\hphantom{0}38.3 \pm 5.0$ & $\hphantom{0}41.7 \pm 4.0$ & $\hphantom{0}43.4 \pm 2.9$ & $\hphantom{0}43.9 \pm 3.1$ & $\hphantom{0}47.9 \pm 3.0$ \\
    {\small \texttt{psychlab continuous recognition}} & $\hphantom{0}52.2 \pm 0.3$ & $\hphantom{0}52.2 \pm 0.3$ & $\hphantom{0}52.2 \pm 0.2$ & $\hphantom{0}52.1 \pm 0.3$ & $\hphantom{0}52.1 \pm 0.2$ \\
    {\small \texttt{psychlab sequential comparison}} & $\hphantom{0}75.6 \pm 0.6$ & $\hphantom{0}76.0 \pm 0.3$ & $\hphantom{0}75.9 \pm 0.9$ & $\hphantom{0}75.8 \pm 0.4$ & $\hphantom{0}75.7 \pm 0.3$ \\
    {\small \texttt{psychlab visual search}} & $101.2 \pm 0.4$ & $101.3 \pm 0.5$ & $100.3 \pm 1.0$ & $101.1 \pm 0.2$ & $101.0 \pm 0.1$ \\
    {\small \texttt{rooms collect good objects test}} & $\hphantom{0}95.0 \pm 1.0$ & $\hphantom{0}96.6 \pm 0.3$ & $\hphantom{0}95.7 \pm 0.3$ & $\hphantom{0}96.5 \pm 0.3$ & $\hphantom{0}96.6 \pm 0.2$ \\
    {\small \texttt{rooms exploit deferred effects test}} & $\hphantom{0}36.9 \pm 1.2$ & $\hphantom{0}36.4 \pm 1.1$ & $\hphantom{0}37.7 \pm 0.6$ & $\hphantom{0}37.3 \pm 0.8$ & $\hphantom{0}37.7 \pm 0.6$ \\
    {\small \texttt{rooms keys doors puzzle}} & $\hphantom{0}48.6 \pm 1.5$ & $\hphantom{0}49.2 \pm 1.7$ & $\hphantom{0}47.8 \pm 1.9$ & $\hphantom{0}55.1 \pm 1.3$ & $\hphantom{0}56.2 \pm 1.9$ \\
    {\small \texttt{rooms select nonmatching object}} & $\hphantom{0}62.9 \pm 11.1$ & $\hphantom{0}65.3 \pm 9.2$ & $\hphantom{0}70.6 \pm 8.9$ & $\hphantom{0}65.7 \pm 9.2$ & $\hphantom{0}60.8 \pm 5.8$ \\
    {\small \texttt{rooms watermaze}} & $\hphantom{0}42.2 \pm 2.7$ & $\hphantom{0}43.3 \pm 1.6$ & $\hphantom{0}42.1 \pm 1.7$ & $\hphantom{0}45.6 \pm 2.1$ & $\hphantom{0}45.0 \pm 1.4$ \\
    {\small \bf \texttt{skymaze irreversible path hard}} & $\hphantom{0}25.5 \pm 2.2$ & $\hphantom{0}27.5 \pm 1.1$ & $\hphantom{0}24.6 \pm 1.6$ & $\hphantom{0}27.8 \pm 1.1$ & $\mathbf{\hphantom{0}32.2 \pm 1.1}$ \\
    {\small \bf \texttt{skymaze irreversible path varied}} & $\hphantom{0}49.0 \pm 2.0$ & $\hphantom{0}49.2 \pm 2.2$ & $\hphantom{0}45.0 \pm 2.5$ & $\hphantom{0}48.1 \pm 1.5$ & $\mathbf{\hphantom{0}53.3 \pm 1.3}$ \\
  \bottomrule
  \end{tabular}
  \caption{Human normalized scores across tasks in the last $5\%$ of training---the last $500\mathrm{M}$ out of $10\mathrm{B}$ frames.
  Statistically significant performance improvements in bold.}
  \label{tab:finalPerformance}
\end{table*}

\clearpage

\begin{table*}
\subsection{Atari-57}
  \vspace{2em}
  \small
  \centering
  \begin{tabular}{lcccc}
  \toprule
     & \multicolumn{1}{c}{RL only} & \multicolumn{1}{c}{Pixel Control} & \multicolumn{1}{c}{CPC} & \multicolumn{1}{c}{PBL} \\
    \midrule 
    {\small \texttt{alien}} & $\hphantom{0}15.8 \pm 1.3$ & $\hphantom{0}16.3 \pm 1.3$ & $\hphantom{0}17.2 \pm 0.7$ & $\hphantom{0}18.9 \pm 2.0$ \\
    {\small \bf \texttt{amidar}} & $\hphantom{0}\hphantom{0}8.6 \pm 0.9$ & $\hphantom{0}\hphantom{0}9.4 \pm 0.8$ & $\hphantom{0}\hphantom{0}9.5 \pm 0.5$ & $\mathbf{\hphantom{0}12.0 \pm 1.2}$ \\
    {\small \texttt{assault}} & $524.0 \pm 47.5$ & $658.2 \pm 44.2$ & $557.1 \pm 48.4$ & $764.7 \pm 105.7$ \\
    {\small \texttt{asterix}} & $\hphantom{0}29.7 \pm 8.4$ & $\hphantom{0}32.8 \pm 9.1$ & $\hphantom{0}30.9 \pm 4.3$ & $\hphantom{0}43.5 \pm 6.1$ \\
    {\small \texttt{asteroids}} & $\hphantom{0}\hphantom{0}3.6 \pm 0.1$ & $\hphantom{0}\hphantom{0}3.7 \pm 0.1$ & $\hphantom{0}\hphantom{0}3.8 \pm 0.1$ & $\hphantom{0}\hphantom{0}4.1 \pm 0.2$ \\
    {\small \texttt{atlantis}} & $5033.8 \pm 328.2$ & $5180.9 \pm 404.8$ & $4761.6 \pm 362.4$ & $5448.9 \pm 729.6$ \\
    {\small \texttt{bank heist}} & $157.9 \pm 11.2$ & $167.7 \pm 6.3$ & $166.9 \pm 10.1$ & $169.3 \pm 21.3$ \\
    {\small \texttt{battle zone}} & $\hphantom{0}73.1 \pm 5.9$ & $\hphantom{0}78.3 \pm 1.7$ & $\hphantom{0}80.3 \pm 4.9$ & $\hphantom{0}88.5 \pm 8.1$ \\
    {\small \texttt{beam rider}} & $\hphantom{0}13.8 \pm 1.2$ & $\hphantom{0}21.3 \pm 1.1$ & $\hphantom{0}13.0 \pm 0.6$ & $\hphantom{0}19.2 \pm 1.8$ \\
    {\small \texttt{berzerk}} & $\hphantom{0}14.6 \pm 1.0$ & $\hphantom{0}15.4 \pm 0.9$ & $\hphantom{0}14.8 \pm 0.7$ & $\hphantom{0}16.2 \pm 1.6$ \\
    {\small \texttt{bowling}} & $\hphantom{0}\hphantom{0}3.4 \pm 1.4$ & $\hphantom{0}\hphantom{0}4.8 \pm 2.0$ & $\hphantom{0}\hphantom{0}5.4 \pm 1.2$ & $\hphantom{0}\hphantom{0}7.3 \pm 1.5$ \\
    {\small \texttt{boxing}} & $781.9 \pm 14.1$ & $765.9 \pm 45.2$ & $793.7 \pm 8.5$ & $807.3 \pm 2.0$ \\
    {\small \bf \texttt{breakout}} & $934.1 \pm 68.6$ & $1034.1 \pm 58.1$ & $997.5 \pm 43.2$ & $\mathbf{1203.0 \pm 68.0}$ \\
    {\small \bf \texttt{centipede}} & $\hphantom{0}-7.6 \pm 0.8$ & $\hphantom{0}-8.5 \pm 0.8$ & $\hphantom{0}-6.8 \pm 0.9$ & $\mathbf{\hphantom{0}-2.6 \pm 1.8}$ \\
    {\small \texttt{chopper command}} & $105.0 \pm 6.2$ & $107.7 \pm 7.7$ & $111.2 \pm 6.6$ & $114.5 \pm 13.7$ \\
    {\small \texttt{crazy climber}} & $280.5 \pm 15.8$ & $299.9 \pm 12.3$ & $293.4 \pm 13.6$ & $292.0 \pm 29.3$ \\
    {\small \texttt{defender}} & $278.0 \pm 41.5$ & $285.7 \pm 36.9$ & $342.1 \pm 41.9$ & $347.1 \pm 60.4$ \\
    {\small \texttt{demon attack}} & $549.9 \pm 101.4$ & $750.6 \pm 108.1$ & $510.3 \pm 54.0$ & $786.1 \pm 148.7$ \\
    {\small \texttt{double dunk}} & $691.7 \pm 108.6$ & $815.8 \pm 148.7$ & $745.1 \pm 38.8$ & $862.9 \pm 162.2$ \\
    {\small \texttt{enduro}} & $113.9 \pm 11.3$ & $129.5 \pm 4.8$ & $121.9 \pm 8.0$ & $141.8 \pm 14.6$ \\
    {\small \texttt{fishing derby}} & $179.5 \pm 12.1$ & $182.7 \pm 3.9$ & $186.8 \pm 9.8$ & $187.0 \pm 23.0$ \\
    {\small \texttt{freeway}} & $103.4 \pm 3.5$ & $108.7 \pm 0.6$ & $107.5 \pm 1.8$ & $102.8 \pm 12.6$ \\
    {\small \texttt{frostbite}} & $\hphantom{0}\hphantom{0}4.4 \pm 0.1$ & $\hphantom{0}\hphantom{0}4.7 \pm 0.2$ & $\hphantom{0}\hphantom{0}4.6 \pm 0.2$ & $\hphantom{0}\hphantom{0}4.6 \pm 0.3$ \\
    {\small \texttt{gopher}} & $935.0 \pm 156.1$ & $989.4 \pm 277.7$ & $838.7 \pm 142.4$ & $1080.0 \pm 235.3$ \\
    {\small \bf \texttt{gravitar}} & $\hphantom{0}44.7 \pm 3.7$ & $\hphantom{0}48.0 \pm 3.7$ & $\hphantom{0}47.5 \pm 4.1$ & $\mathbf{\hphantom{0}54.6 \pm 1.0}$ \\
    {\small \texttt{hero}} & $\hphantom{0}37.1 \pm 1.9$ & $\hphantom{0}40.1 \pm 0.8$ & $\hphantom{0}39.2 \pm 1.0$ & $\hphantom{0}38.8 \pm 4.6$ \\
    {\small \texttt{ice hockey}} & $149.7 \pm 6.0$ & $159.2 \pm 5.8$ & $152.9 \pm 4.6$ & $158.6 \pm 18.0$ \\
    {\small \texttt{jamesbond}} & $141.2 \pm 13.0$ & $166.4 \pm 11.3$ & $153.4 \pm 10.0$ & $155.4 \pm 28.1$ \\
    {\small \texttt{kangaroo}} & $\hphantom{0}50.5 \pm 2.6$ & $\hphantom{0}53.2 \pm 1.4$ & $\hphantom{0}67.2 \pm 27.7$ & $112.9 \pm 42.3$ \\
    {\small \texttt{krull}} & $669.1 \pm 26.4$ & $704.9 \pm 14.4$ & $698.0 \pm 9.3$ & $719.7 \pm 42.8$ \\
    {\small \bf \texttt{kung fu master}} & $104.0 \pm 6.7$ & $112.4 \pm 7.4$ & $112.1 \pm 9.4$ & $\mathbf{150.5 \pm 17.4}$ \\
    {\small \texttt{montezuma revenge}} & $\hphantom{0}\hphantom{0}0.0 \pm 0.0$ & $\hphantom{0}\hphantom{0}0.0 \pm 0.0$ & $\hphantom{0}\hphantom{0}0.0 \pm 0.0$ & $\hphantom{0}\hphantom{0}0.0 \pm 0.0$ \\
    {\small \texttt{ms pacman}} & $\hphantom{0}24.2 \pm 1.2$ & $\hphantom{0}25.7 \pm 1.0$ & $\hphantom{0}26.8 \pm 0.8$ & $\hphantom{0}28.2 \pm 3.3$ \\
    {\small \texttt{name this game}} & $137.7 \pm 13.2$ & $153.7 \pm 17.9$ & $110.4 \pm 17.6$ & $165.9 \pm 10.2$ \\
    {\small \texttt{phoenix}} & $111.9 \pm 8.9$ & $134.7 \pm 14.6$ & $124.1 \pm 7.5$ & $143.3 \pm 13.7$ \\
    {\small \texttt{pitfall}} & $\hphantom{0}\hphantom{0}3.2 \pm 0.1$ & $\hphantom{0}\hphantom{0}3.3 \pm 0.1$ & $\hphantom{0}\hphantom{0}3.3 \pm 0.0$ & $\hphantom{0}\hphantom{0}3.3 \pm 0.1$ \\
    {\small \texttt{pong}} & $108.2 \pm 5.1$ & $109.7 \pm 8.2$ & $114.5 \pm 2.5$ & $116.9 \pm 0.4$ \\
    {\small \texttt{private eye}} & $\hphantom{0}\hphantom{0}0.1 \pm 0.0$ & $\hphantom{0}\hphantom{0}0.1 \pm 0.0$ & $\hphantom{0}\hphantom{0}0.1 \pm 0.0$ & $\hphantom{0}\hphantom{0}0.1 \pm 0.0$ \\
    {\small \texttt{qbert}} & $\hphantom{0}37.9 \pm 11.5$ & $\hphantom{0}47.9 \pm 15.2$ & $\hphantom{0}16.4 \pm 6.5$ & $\hphantom{0}72.8 \pm 20.3$ \\
    {\small \texttt{riverraid}} & $\hphantom{0}27.4 \pm 5.5$ & $\hphantom{0}37.3 \pm 5.3$ & $\hphantom{0}18.9 \pm 6.5$ & $\hphantom{0}23.0 \pm 8.7$ \\
    {\small \texttt{road runner}} & $321.0 \pm 40.2$ & $382.0 \pm 44.2$ & $359.4 \pm 57.5$ & $460.2 \pm 81.3$ \\
    {\small \texttt{robotank}} & $218.8 \pm 24.6$ & $282.5 \pm 13.6$ & $207.6 \pm 9.2$ & $274.1 \pm 32.6$ \\
    {\small \texttt{seaquest}} & $\hphantom{0}\hphantom{0}5.0 \pm 0.4$ & $\hphantom{0}\hphantom{0}6.3 \pm 1.5$ & $\hphantom{0}\hphantom{0}5.8 \pm 0.2$ & $\hphantom{0}\hphantom{0}8.9 \pm 2.6$ \\
    {\small \texttt{skiing}} & $\hphantom{0}17.4 \pm 2.5$ & $\hphantom{0}18.1 \pm 1.9$ & $\hphantom{0}13.5 \pm 2.7$ & $\hphantom{0}21.0 \pm 2.9$ \\
    {\small \texttt{solaris}} & $\hphantom{0}\hphantom{0}8.9 \pm 0.9$ & $\hphantom{0}\hphantom{0}8.0 \pm 0.7$ & $\hphantom{0}10.0 \pm 1.1$ & $\hphantom{0}\hphantom{0}9.1 \pm 0.9$ \\
    {\small \bf \texttt{space invaders}} & $\hphantom{0}40.2 \pm 3.9$ & $\hphantom{0}48.3 \pm 6.8$ & $\hphantom{0}41.3 \pm 2.9$ & $\mathbf{106.1 \pm 13.0}$ \\
    {\small \texttt{star gunner}} & $377.1 \pm 37.3$ & $436.4 \pm 47.7$ & $415.0 \pm 37.9$ & $477.5 \pm 62.8$ \\
    {\small \texttt{surround}} & $\hphantom{0}73.9 \pm 7.7$ & $\hphantom{0}86.8 \pm 10.2$ & $\hphantom{0}83.4 \pm 8.8$ & $\hphantom{0}94.2 \pm 12.4$ \\
    {\small \texttt{tennis}} & $287.6 \pm 10.7$ & $275.1 \pm 28.9$ & $255.3 \pm 36.8$ & $231.4 \pm 48.2$ \\
    {\small \texttt{time pilot}} & $849.3 \pm 59.3$ & $910.9 \pm 79.5$ & $783.6 \pm 71.0$ & $999.7 \pm 148.4$ \\
    {\small \texttt{tutankham}} & $144.1 \pm 6.4$ & $148.0 \pm 6.5$ & $148.8 \pm 5.0$ & $154.3 \pm 13.2$ \\
    {\small \texttt{up n down}} & $2388.1 \pm 238.8$ & $2573.1 \pm 150.9$ & $2579.7 \pm 142.9$ & $2832.4 \pm 284.6$ \\
    {\small \texttt{venture}} & $\hphantom{0}\hphantom{0}0.0 \pm 0.0$ & $\hphantom{0}\hphantom{0}0.0 \pm 0.0$ & $\hphantom{0}\hphantom{0}0.0 \pm 0.0$ & $\hphantom{0}\hphantom{0}0.0 \pm 0.0$ \\
    {\small \texttt{video pinball}} & $1139.6 \pm 101.6$ & $1121.1 \pm 79.4$ & $1072.3 \pm 109.4$ & $1234.1 \pm 185.6$ \\
    {\small \texttt{wizard of wor}} & $\hphantom{0}93.7 \pm 10.0$ & $107.6 \pm 6.9$ & $\hphantom{0}96.7 \pm 6.5$ & $108.5 \pm 14.6$ \\
    {\small \bf \texttt{yars revenge}} & $\hphantom{0}19.3 \pm 2.3$ & $\hphantom{0}36.1 \pm 14.3$ & $\hphantom{0}24.7 \pm 2.4$ & $\mathbf{\hphantom{0}83.8 \pm 17.9}$ \\
    {\small \texttt{zaxxon}} & $136.5 \pm 12.2$ & $176.0 \pm 10.9$ & $159.3 \pm 7.9$ & $169.4 \pm 27.9$ \\

  \bottomrule
  \end{tabular}
  \caption{Human normalized scores across tasks in the last $5\%$ of training---the last $500\mathrm{M}$ out of $10\mathrm{B}$ frames. 
  Statistically significant performance improvements in bold.}
  \label{tab:finalAtariPerformance}
  \normalsize
\end{table*}

\end{document}